\pdfoutput=1
\documentclass[11pt]{article}
\usepackage[final]{acl}
\usepackage{times}
\usepackage{latexsym}
\usepackage[T1]{fontenc}
\usepackage[utf8]{inputenc}
\usepackage{microtype}
\usepackage{inconsolata}
\usepackage{graphicx}
\usepackage[most]{tcolorbox}
\definecolor{lightblue}{RGB}{230, 240, 255}
\definecolor{softpurple}{RGB}{240, 230, 255}
\definecolor{darkblue}{RGB}{80, 120, 180}  
\definecolor{pastelblue}{RGB}{225, 240, 255}
\usepackage{caption}
\usepackage{comment}
\usepackage{amsmath}
\usepackage{todonotes}
\usepackage{array}
\usepackage{booktabs}
\usepackage{enumitem} 
\usepackage{tikz}
\usepackage{float}
\begin{document}

\title{LLMs and their Limited Theory of Mind: \\Evaluating Mental State Annotations in Situated Dialogue}

\author{
    Katharine Kowalyshyn \and Matthias Scheutz \\
    Department of Computer Science \\
    Tufts University \\
    \texttt{\{Katharine.Kowalyshyn, Matthias.Scheutz\}@tufts.edu} \\ 
}

\maketitle

\begin{abstract}
Effective human teams excel at maintaining a consistent {\em shared
mental model} (SMM) that reflects the shared understanding of
individual team members about the task and what remains to be done.
We present a novel, two-step framework that leverages large language
models (LLMs) both as (1) generators of mental model traces from team dialogues to track the team's SMM and (2) automated detectors of discrepancies between inferred and ground truth mental model traces. We define an SMM coherence evaluation framework for
this use case and apply it to six dialogues in a previously published
team corpus, ultimately producing a dataset of human and LLM SMM
mental model traces, a reproducible evaluation framework for SMM coherence,
and an empirical assessment of LLM-based discrepancy detection. Our
results reveal that while LLMs exhibit apparent coherence on
straightforward natural-language trace generation tasks, they systematically
err in scenarios requiring spatial reasoning or disambiguation of
transcription-level disfluencies.

\end{abstract}
 
\section{Introduction}

\begin{figure}[h]
  \centering
  \includegraphics[width=0.85\columnwidth, keepaspectratio]{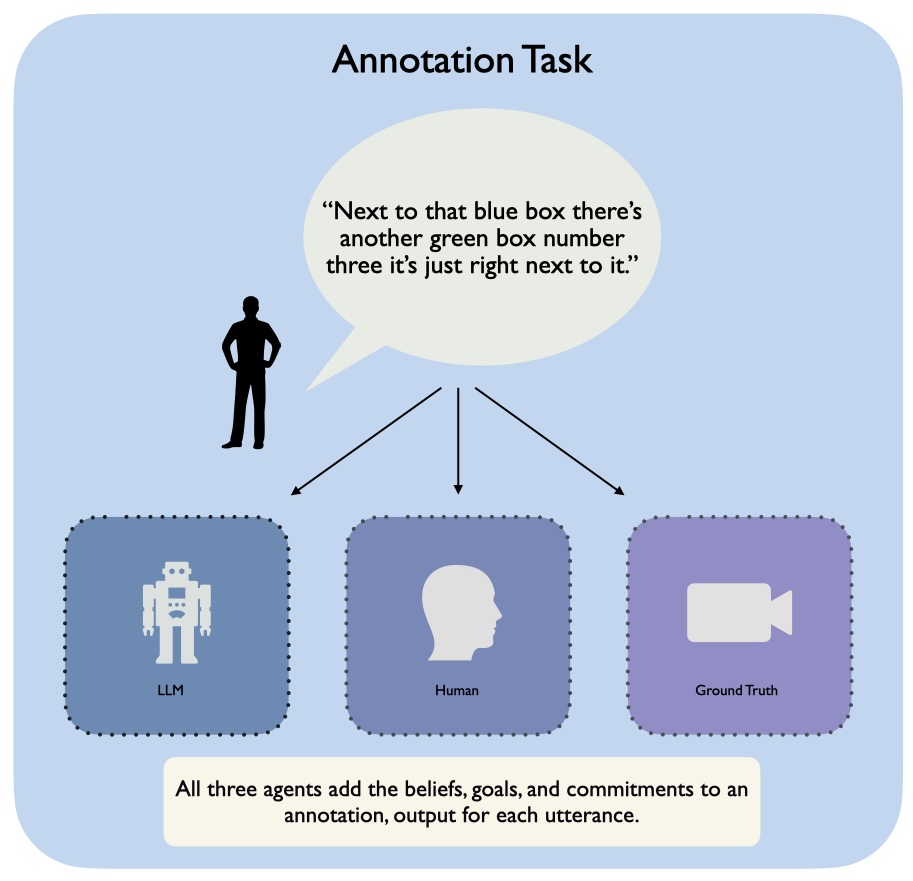}
  \caption{Example of the first stage of our evaluation pipeline. A snippet of situated team dialogue is independently processed by an LLM, a naive human, and a video-informed human with access to the environment, each producing a mental model trace of belief, commitment, and goal states for each agent. The audio-only traces (LLM and naive human) are later compared against the ground truth trace by an LLM judge to assess ToM-like reasoning capabilities.}
  \label{fig1}
\end{figure}

Successful team performance often hinges not only on what is said, but
on what is understood, including the latent belief states, goals, and
commitments shared among team members. In tasks which require
coordination through spoken communication, situated dialogue, embedded
in real-time decision-making and environmental context, becomes the
primary conduit for this shared understanding called {\em shared
mental model} (e.g., \cite{scheutz2017framework}). For humans, the
subtext behind task-based communications (i.e., the assumptions,
intentions, and knowledge of others) is often grasped implicitly and
the question we ask is whether large language models (LLMs) could do
the same.  More precisely, can LLMs accurately infer and thus annotate
internal belief states of team members based on dialogue interactions
and detect discrepancies in their mental representations?

\begin{figure*}
  \centering
  \includegraphics[%
    width=0.85\textwidth,
    keepaspectratio%
  ]{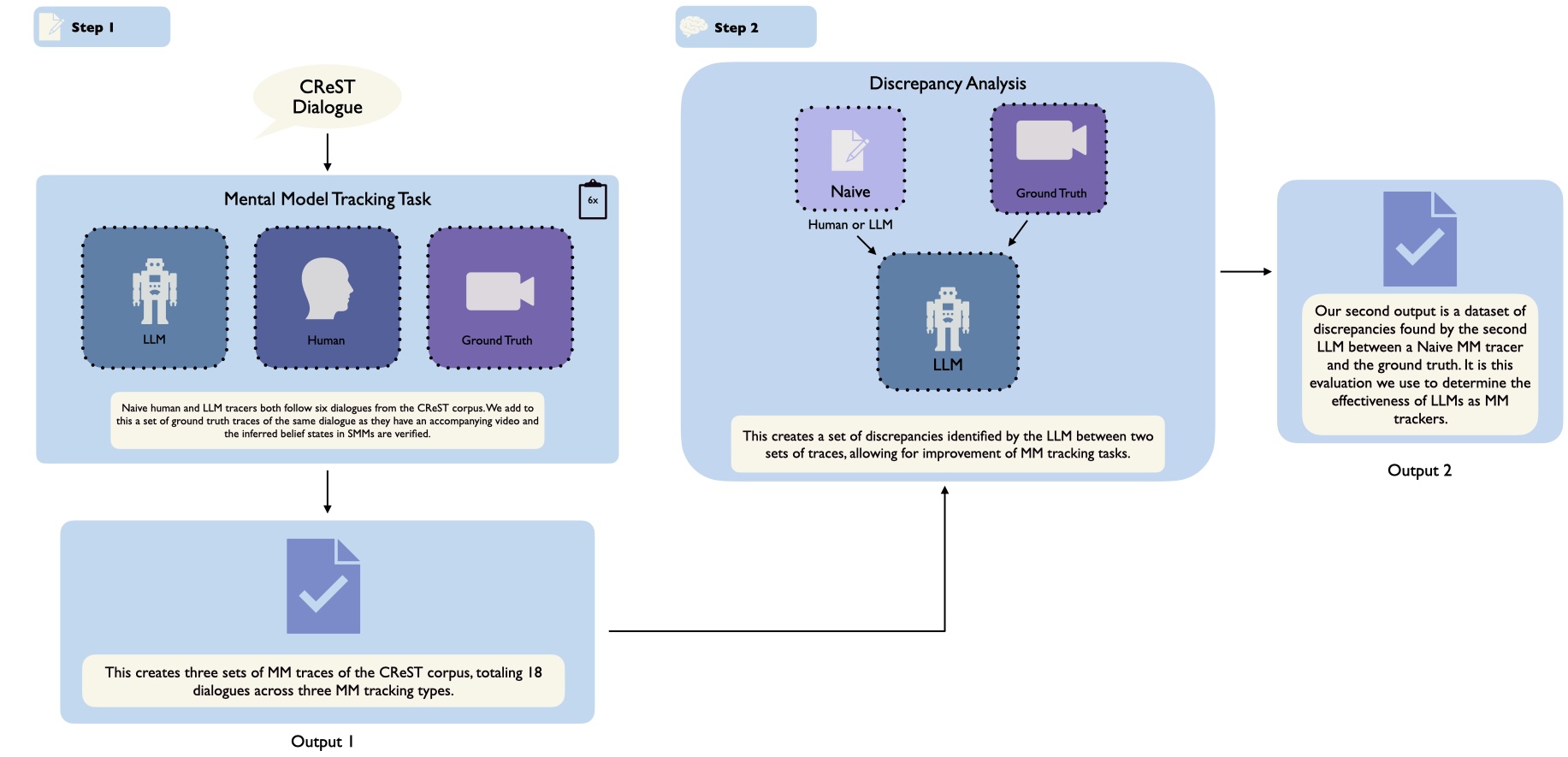}
  \caption{An overview of our two-step process used to trace and find discrepancies between SMMs using LLMs. We are able to create two outputs: (1) a dataset of ground truth, human, and LLM mental model traces and (2) a dataset of discrepancies between ground truth traces and human/LLM mental model traces. }
  \label{fig:overview}
\end{figure*}

In this paper, we treat situated dialogue as a proxy task for
assessing {\em Theory of Mind} (ToM) capabilities in current large
language models, focusing on team coherence in situated dialogue.  We
do so by utilizing the {\em Cooperative Remote Search Task} (CReST)
corpus \cite{eberhard} which consists of situated task-based
interactions between a local \textit{searcher} and
remote \textit{director}.  As part of our experimental setup, we
evaluate three independent LLMs, OpenAI’s o3-mini, Anthropic’s Claude
Sonnet 4, and Google’s Gemma, alongside a pair of naive humans and
another pair of humans with access to the ground truth, to each
generate mental model traces for six CReST dialogues for which video recordings are available
to establish ground truth traces \footnote{3 LLMs × 6 dialogues + 1 naive human pair × 6 dialogues + 1 ground truth human pair × 6 dialogues = 30 total trace sets. The 24 comparisons derive from comparing each of the 4 non-ground-truth trace generators across 6 dialogues.}. We then employed a separate ``LLM judge'' to compare the audio-only traces (both human and LLM) against the ground truth traces, which we call ``discrepancy analysis''. Additionally, we outline a discrepancy evaluation framework which can apply to many team-based coordination tasks, extending the utility of this paper beyond the empirical results we provide.

This two-step approach (as seen in Figure~\ref{fig:overview}) yields a dataset of mental model traces, a reproducible evaluation pipeline defined by fixed prompts, tracing procedures, discrepancy metrics, and an empirical assessment of LLM discrepancy detection. While our primary goal is to assess LLMs'
ability to trace belief states in team dialogue, our findings also
underscore a broader point: LLMs simulate limited but measurable
ToM-like behaviors in situated contexts. To our knowledge, this is the
first study to use belief-state traces in real-world dialogue as
a lens for evaluating ToM in LLMs. As such, we propose situated
dialogue trace generation as a viable proxy task for probing ToM
capabilities in AI systems. The remainder of this paper is structured as follows: Section 2 reviews related work on LLMs as generators of mental models and team coherence. Section 3 describes the CReST corpus. Section 4 introduces our discrepancy framework. Section 5 presents results, and Section 6 discusses implications and limitations.

\section{Related Work}
\label{relatedwork}

LLMs have shown impressive capabilities for natural language
understanding and for performing novel tasks such as interpreting
natural language expressions and mapping them into new formalisms,
from programs, to rules, to logics
\cite{pan2023logic, manasetal, zhou2024don}.
While recent work has explored ToM in LLMs, many existing evaluations
focus on static, decontextualized tasks such as multiple-choice
questions, short story comprehension, or isolated inference
problems. These tasks often test factual or scripted understanding
rather than the dynamic, real-time reasoning that underlies ToM in
human communication \cite{chen2024tombenchbenchmarkingtheorymind,
Kosinski_2024, strachan2024testing, gandhi2023understanding}. In
contrast, situated dialogue involves ongoing belief tracking,
contextual inference, and implicit coordination, which are abilities
more reflective of how ToM functions in natural human
interaction. Recently there has even been work negating the idea that
so-called thinking/reasoning models can reason at
all \cite{illusion-of-thinking}. This highlights a critical gap in the
current literature: few, if any, ToM evaluations challenge models to
reason about mental states as they unfold in interactive, grounded
team contexts.  Here we briefly review recent work on using LLMs
for tracking human mental states in task-based dialogues.

\subsection{LLMs for Mental State Inference}
 
LLMs have rapidly become integral to modern corpus construction and
annotation, spanning direct label suggestion, iterative human–LLM
editing, and conversational scaffolding \cite{Yu_Li_Su_Fuoli_2024,
zhu2025multimodal, Wu_Mu_Zhou_Ma_Chen_Liu_2024, zhao2024decoratelm,
Weissweiler_Koksal_Schutze_2024, Pangakis_Wolken_2024}. Together,
these approaches reduce human workload while improving annotation
consistency and scalability across a wide range of NLP tasks. Still, no prior work combines LLM-generated mental model traces with the inference of mental states in team dialogue.
 
 \subsection{LLMs for Team Coherence}
 Recent work in multi-party conversation tracking, role-playing, and
peer agents all include LLMs as a main character in the task
\cite{Addlesee,Kong_Zhao_Chen_Li_Qin_Sun_Zhou_Zhou_Sun_2024,Liu_Yao_An_Wang_2024, palmerspeech},
but LLM reasoning abilities have repeatedly been questioned despite
their benchmark performance
\cite{stechly2024selfverificationlimitationslargelanguage,
  mondorf2024accuracyevaluatingreasoningbehavior,
  bertolazzi2024systematicanalysislargelanguage}.  While some hypothesize
that ToM reasoning has already emerged as a side effect of
training data \cite{ma2023towards,
  Kosinski_2024,Kim_Sclar_Zhou_Bras_Kim_Choi_Sap_2023}, others argue
that, overall, experiments where LLMs have performed well on ToM tasks are not enough evidence for the emergence of cognitive abilities akin to humans
\cite{sarıtas2025systematicreviewevaluationlarge, Ullman_2023}.

In the teaming context, the notion of a {\em shared mental model} began
alongside cognitive science's inception \cite{JOHNSONLAIRD198071}, and
various psychological studies have been performed to understand team
performance in the past, including to generate natural dialogue and
evaluate shared mental models \cite{mathieu2000influence,
  gervits2016team, jonker2010shared}.  Extending this work to
human-AI teams research has included evaluations of
SMMs and their impact on team performance
\cite{andrews2023role, schelble2022let, sarkar2025understandingcommongroundmisalignment}, including human-robot team
performance, establishing that task tracking through SMMs enhances
team performance \cite{scheutz2024multi, edgar2024toward}.  Recent
team coherence research also includes the creation of collaborative
teams of LLMs and LLM understanding of utterances with uncertainty
\cite{Liu_Zhang_Li_Liu_Yang_2024,
  Paige_Soubki_Murzaku_Rambow_Brennan_2024}. Adjacent work on dialogue state tracking and commitment recognition addresses related inference problems but typically focuses on surface task states rather than the full mental model of each agent, including second-order beliefs \cite{sarkar2025understandingcommongroundmisalignment}. Parallel annotation work on 
asymmetric dialogue has proposed perspectivist schemes for capturing grounded 
misunderstandings between agents with unequal environmental access~\cite{li2026groundedmisunderstandingsasymmetricdialogue}, 
directly motivating the audio-only versus video-grounded distinction central to our 
design.

Despite rich work on human SMMs and on LLM mental state inference in isolation, no
prior work evaluates LLMs' ability to infer mental states in
team-based dialogues.  Overall, given that SMMs are useful for human
teams, that humans are able to build and update their mental
models of other team members through dialogues, and that there is
an increasing interest in mixed-initiative human-AI teams, the
question addressed in this paper is timely -- to our knowledge there
is no other work that investigates the ability of LLMs to make mental
state inferences from task-based dialogues and compares them to human
trace generation performance.

\section{Dataset: The CReST Corpus}

To investigate the ability of LLMs for dialogue-based inference about
human mental states, we utilized the ``Cooperative Remote Search
Task'' (CReST) corpus which is comprised of task-based dialogues from
23 human dyad teams and was originally created to fill the void of a
lack of naturalistic dialogue in team-based
settings \cite{eberhard}. Each group consisted of one person
designated to be a searcher who was present in an environment, while
the other person, the director, was not physically present but was
given a map of the space which was intentionally incomplete (full map
is shown in Appendix~\ref{sec:extra}). The pair communicated via audio
channel and were given a series of tasks to complete. Tasks included
locating boxes in the environment, labeling boxes on the map, and
changing the locations of blocks which were found within
specifically-colored boxes.

The resulting corpus contains 40,083 words in 5,872 sentences between
all twenty-three pairs \cite{kubler2012parallel, eberhard, gervits2016team}. Crucial
to this paper, this corpus highlights disfluencies and typical speech
patterns, which allowed us to conduct a more natural simulation of SMM
tracking in real time situated dialogues.  Here, we use the
transcribed text from six dialogues to generate mental model traces for which
additional video recordings were available from the searcher's headcam
(Video image in Appendix~\ref{sec:extra}), which allowed us to
determine the ground truth of where the searcher was located and what
the searcher was seeing at all times during the task-based
interactions (for the stationary director this was not necessary).  We
also generate traces from the same dialogues using naive annotators
sans access to the videos, to explore the agreement rate between human
annotators and LLMs. The following sections outline the
mental model tracking procedure for both naive human and LLM, and ground truth
inference-based traces.

\subsection{Mental Model Tracking Procedure}

All of the agents' mental states were first traced by two humans across all dialogues. Throughout this section, we use ``annotator'' to refer specifically to the human workers hired to label the dialogues; the outputs they produce are mental model traces in the framework described in Section 4. These human annotators were undergraduate students
hired and compensated at an hourly rate of \$17 in accordance with
university policy. Prior to annotation, both annotators received a training session in which the task 
structure, annotation schema, and field definitions were explained using worked examples 
drawn from outside the six target dialogues. Annotations were produced in free-form 
natural language rather than from a fixed candidate list, with the constraint that 
belief, goal, and commitment fields had to use the verb vocabulary specified in the 
LLM prompt (e.g., \textit{at}, \textit{in}, \textit{near}) to ensure comparability 
with LLM-generated traces. The concept of ``commitment'' was operationalized for 
annotators as an agent's explicit agreement to take a specific action, distinguishable 
from a goal (desired outcome) by the presence of a verbal or behavioral uptake signal 
in the dialogue. Annotators were encouraged to ask clarifying questions during training 
but worked independently during annotation.

\begin{figure}
	\centering
    \includegraphics[width=0.9\columnwidth]{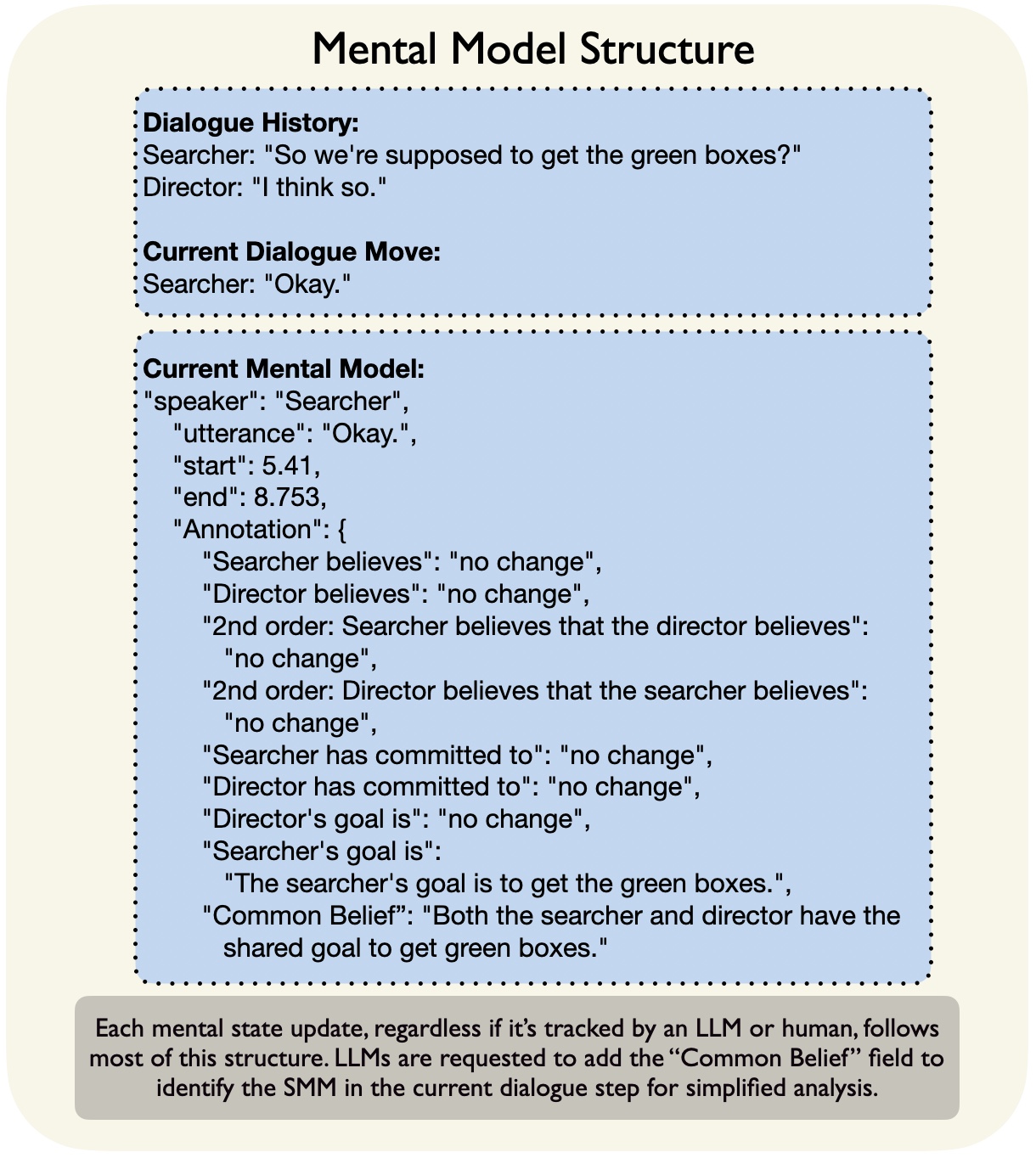} \caption{Using an example from Dialogue 1, we show the trace structure used by LLMs to identify shared mental models and the common beliefs held by agents. Our dataset of traces includes a mental state update of this structure for each utterance.}  \label{fig:annstructure} \vspace{-5mm}
\end{figure}

We hired two annotators to first naively annotate
(without ground truth videos) the dialogues for mental states
independently. Then, the two annotators sat together and resolved any
disagreements, creating a dataset of agreed upon naive annotations
with an agreement rate of 100\% post-adjudication. Following this, we completed the same procedure but with the
ground truth videos. See Figure~\ref{fig:annstructure} for the
annotation format. In total, we had six dialogues consisting
of \textbf{1,142} utterances and a resulting two-part dataset with
ground truth and naive annotations from humans (see
Table~\ref{tab:utterances} below for more detail).

\begin{table}[h]
  \centering
  \scriptsize
  \resizebox{0.75\columnwidth}{!}{
  \begin{tabular}{cc} 
    \toprule
    Dialogue & Number of Utterances  \\
    \midrule
    1 & 173 \\
    2 & 147 \\
    3 & 225 \\
    4 & 162 \\
    5 & 241 \\
    6 & 194 \\
    \textbf{Total} & \textbf{1,142} \\
    \bottomrule
  \end{tabular}%
  }
  \caption{Total utterances for each dialogue in our dataset created from the CReST corpus. This includes all of the dialogues from the corpus which have a corresponding video, which is necessary in order to establish ground truth mental state traces.}
  \label{tab:utterances}
\end{table}

We generated our dataset (Step 1 in Figure~\ref{fig:overview}) by
having three groups of trace generators label the same six dialogues for the
searcher's and director's beliefs, goals, and commitments.  The first
set of trace generators was a series of naive LLMs which were naive as they
were only given limited information about the environment: prompts
detailed task overview (see Appendix~\ref{sec:full_prompts} for
prompts) but there was no plausible way for the LLM to establish a
ground truth understanding of the environment operating on natural
language interaction alone; all updates about the state of the
environment had to be inferred from the utterances.  The second set of
mental state trackers were two ``naive humans'' who similarly only had the
language dialogues available and were compensated and completed the
adjudication process as outlined above. Figure~\ref{fig:annstructure}
shows the annotation structure for each utterance, which was employed
by both human and LLM mental state tracers.

For the third trace generation condition, we employed two humans to generate mental state traces
in the same way, but this time with access to a video which provides
ground truth environmental state from the searcher's point-of-view
recorded from a headcam as well as the director's screen with the map,
and audio between the two. This ground-truth traced mental state set, corroborated by
two humans, served as our ground truth comparison for the LLMs' (and the
naive humans') mental trace performance.

The key strength of our approach is the use of video recordings to
establish ground truth mental model traces. The use of video recordings to establish ground truth traces, and the distinction between audio-only and video-grounded inference, are discussed in detail in Section 4.

Following the creation of our dataset, we were able to move on to Step
2 of Figure~\ref{fig:overview}, the tracing discrepancy analysis.

\section{Discrepancy Analysis Framework}
\label{framework}

There are two separate applications of LLMs in this work: one where an
LLM generates mental model traces for the dialogue agents, and a second
where a separate LLM judge identifies discrepancies between those traces
and the ground truth traces derived from video. For the latter, the
question arises how to detect and measure discrepancies between an
inferred mental model trace and the ground truth trace. We therefore
first introduce our discrepancy framework and metric, then apply it to
our dataset.

It is important to distinguish two sources of divergence between an
audio-only trace and the ground truth trace:

\begin{enumerate}[noitemsep]
    \item \textbf{Trace divergence due to missing perceptual information:}
    cases where the ground truth trace contains a mental model update
    that cannot be inferred from the transcript alone, because it
    derives from a perception or action not verbalized by either agent.
    \item \textbf{Mental model discrepancies:} cases where the
    audio-only trace assigns a different belief, goal, or commitment
    than the ground truth trace even for fields that \textit{could} in
    principle be inferred from the transcript.
\end{enumerate}

These two sources of divergence are related: trace divergence implies
mental model discrepancies. However, the converse does not hold:
discrepancies in individual mental model fields may arise from
inferential errors rather than missing perceptual information. Our
discrepancy framework operates at the level of individual field
comparisons between the audio-only and ground truth traces.

\subsection{Discrepancy Types}
\label{discframe}

The four discrepancy types defined below were derived inductively through a pilot 
annotation phase conducted on two dialogues outside the six used in our main 
evaluation. During piloting, two of the authors independently reviewed mismatches 
between audio-only and ground truth traces and grouped them by the nature of the 
error. The four categories emerged as exhaustive and mutually exclusive with respect 
to the pilot data; edge cases were resolved by consensus. We acknowledge that 
alternative taxonomies are possible. For instance, omissions and unsupported beliefs 
could be collapsed into a single ``mismatch'' category, and belief 
contradictions could be further subdivided by whether the conflicting proposition is 
spatially or temporally grounded. We adopt the four-way split because it preserves 
theoretically meaningful distinctions that are relevant to 
downstream team coordination applications.

We define a \textit{discrepancy instance} as a mismatch between a
single field in an audio-only mental model state and the corresponding
field in the ground truth state at the same utterance. Each field is
evaluated independently, so discrepancies are counted at the field
level. Our framework identifies four types of discrepancies, which we
now define using a running example. Let $M^*_t(x)$ denote the ``ground
truth'' mental model state of person $x$ at utterance $t$ derived from
video, and let $M_t(x)$ denote the mental model state for person $x$
at the same utterance derived from audio only.  Each mental model
consists of a set of propositions over agents and world states.

Consider then the two mental models after utterance $t$:

\[
M^*_t(\text{searcher}) = \{\texttt{pink\_box}(r_3),\,
\texttt{at}(\text{searcher}, r_3)\}
\]
\[
M_t(\text{searcher}) = \{\texttt{pink\_box}(r_3),\,
\texttt{at}(\text{searcher}, r_2)\}
\]

Here the audio-only trace assigns the searcher to room~2 while the
ground truth places them in room~3. This mismatch constitutes a
discrepancy instance in the \texttt{at(searcher, $\cdot$)} field. The
four discrepancy types below classify the nature of such mismatches.

\subsubsection{Belief Contradictions}


A \textbf{belief contradiction} occurs when, at the same utterance, the searcher and director hold opposing beliefs about the same world state. Using a similar example as above, a belief contradiction would be:

\[
M_t(\text{searcher}) \ni \texttt{at}(\text{green box}, r_3), 
\]
\[
M_t(\text{director}) \ni \texttt{at}(\text{green box}, r_2)
\]

These are mutually exclusive propositions about the green box's
location, and the co-occurrence across the two traces constitutes a
belief contradiction.

\subsubsection{Omissions}


An \textbf{omission} occurs when $M^*_t$ contains a proposition $p$
for a given agent that is absent from $M_t$, i.e., the
audio-only trace simply does not record the corresponding belief or
goal update. For example, if the ground truth trace records that the
director's goal has shifted to directing the searcher toward corridor~2,
but the audio-only trace records \textit{no change} for the director's
goal field at that utterance, this constitutes an omission. 

\subsubsection{Unsupported Beliefs}

An \textbf{unsupported belief} occurs when $M_t$ contains a proposition
$p$ for a given agent that is neither confirmed nor contradicted by
$M^*_t$. It cannot be validated or negated using the ground truth.
These often arise when an LLM infers a belief that is linguistically
plausible but not grounded in any observable dialogue event or
environmental fact. For example, if the audio-only trace records that
the searcher's goal is to follow the director's directions at an
utterance where the ground truth records no goal update, this is an
unsupported belief. Unlike omissions, unsupported beliefs reflect
additions to the mental model trace that have no evidential basis, and
thus represent a distinct failure mode in which the inferred trace
overgenerates.

\subsubsection{False Beliefs}

A \textbf{false belief} discrepancy occurs when an agent's belief disagrees with the ground truth trace, regardless of whether it contradicts the other agent. Suppose $M_t$ contains a
proposition $p$ for a given agent and $M^*_t$ contains a proposition
$q \neq p$ for that same agent, where $p$ and $q$ are not directly
negations of one another (which would constitute a contradiction), but
$p$ is nonetheless incorrect given the ground truth. For example:

\[
M^*_t(\text{director}) \ni \texttt{pink\_box}(r_3),
\]
\[
M_t(\text{director}) \ni \texttt{green\_box}(r_3)
\]

Here the audio-only trace assigns an incorrect object type to room~3 in
the director's mental model state. The ground truth does not assert the
negation of \texttt{green\_box}($r_3$) directly, but the proposition is
nonetheless wrong given the video evidence.

\subsection{Severity of Discrepancies}
\label{severities}

We introduce a severity ranking over the four discrepancy types defined
above. This ranking reflects the degree to which each type of
discrepancy distorts the inferred mental model trace relative to the
ground truth, and is motivated by the downstream impact such
distortions would have on team coordination if acted upon.

Belief contradictions and false belief discrepancies represent direct
mismatches between the propositions assigned to an agent's mental model
state, and therefore most severely distort the inferred trace. An agent
acting on a contradicted or false belief would have a fundamentally
incorrect understanding of the task state. Unsupported beliefs
correspond to spurious additions to the mental model trace that lack
evidential grounding; while not directly harmful in isolation, they
introduce noise into the inferred trace and could mislead downstream
reasoning if acted upon. Omissions are the least severe, as they
represent missing updates rather than incorrect ones, though omitting
critical state updates can still degrade team coherence when the absent
information is task-relevant.

We therefore rank discrepancies in descending order of severity:
\textbf{Belief Contradictions $>$ False Beliefs $>$ Unsupported Beliefs
$>$ Omissions}. In the results reported here, all types are assigned
equal weight as a conservative, task-agnostic baseline; differential
weighting is left to task-specific adaptation. Appendix E provides
per-type, per-utterance breakdowns to support reweighting for
future use cases.

To operationalize this ranking, we define a weighted discrepancy score
per model--dialogue pair $(m, d)$.

Let
\begin{itemize}
    \item $B_{m,d}$ = number of Belief Contradictions
    \item $F_{m,d}$ = number of False Beliefs
    \item $U_{m,d}$ = number of Unsupported Beliefs
    \item $O_{m,d}$ = number of Omissions
    \item $w_{x}$ = weight for respective discrepancy $x$
\end{itemize}

\noindent Then, the weighted raw discrepancy score is

\begin{equation}
r_{m,d} = w_b B_{m,d} + w_f F_{m,d} + w_u U_{m,d} + w_o O_{m,d}
\label{eq:rawscore}
\end{equation}

\noindent To account for dialogue length, we divide by the number of utterances $N_d$ in dialogue $d$, resulting in a per-utterance discrepancy score:

\begin{equation}
s_{m,d} = \frac{r_{m,d}}{N_d}
\label{eq:perutterance}
\end{equation}

\noindent Finally, to enable fair comparisons across models and dialogues, we normalize the scores to the range $[0, 1]$:

\begin{equation}
\mathcal{S}_{m,d} = 1 - \frac{s_{m,d} - s_{\min}}{s_{\max} - s_{\min}}
\label{eq:normalized}
\end{equation}

\noindent Here, $s_{\min}$ and $s_{\max}$ represent the minimum and maximum $s_{m,d}$ values across all $(m, d)$ pairs. A higher $\mathcal{S}_{m,d}$ indicates better alignment with ground truth, i.e., fewer and/or less severe discrepancies per utterance.

This severity-informed metric provides a principled way to compare trace fidelity and model reasoning performance across dialogue complexities.
\label{severities}

\begin{table}
  \centering
  \scriptsize
  \resizebox{\columnwidth}{!}{
  \begin{tabular}{llcccccc}
    \toprule
    \multicolumn{2}{l}{} & \multicolumn{6}{c}{\textbf{Dialogues}} \\
    \multicolumn{2}{l}{} & D1 & D2 & D3 & D4 & D5 & D6 \\
    \midrule

    \multicolumn{2}{l}{\textbf{Annotator}} \\
    & \textbf{o3-mini} \\
    & Belief Contradictions     & 30 & 27 & 51 & 15 & 24 & 34 \\
    & Omissions                 & 55 & 49 & 101 & 52 & 88 & 49 \\
    & Unsupported Beliefs       & 86 &  41 & 79 & 158 & 156 & 125 \\
    & False Beliefs             & 6 & 3 & 3 & 0 & 6 & 0 \\
    & \textbf{Total Discrepancies (o3-mini)} &  204 & \textbf{\textcolor{blue}{120}} & 234 & 225 & 274 & \textbf{\textcolor{blue}{208}} \\
    \addlinespace[1.2ex]

    & \textbf{Claude Sonnet 4} \\
    & Belief Contradictions     & 149 & 139 & 206 & 106 & 178 & 153 \\
    & Omissions                 & 144 & 115 & 226 & 76 & 209 & 160 \\
    & Unsupported Beliefs       & 231 & 179 & 305 & 262 & 303 & 230 \\
    & False Beliefs             & 4 & 2 & 8 & 0 & 7 & 4 \\
    & \textbf{Total Discrepancies (Claude)} & \textbf{\textcolor{violet}{530}} & \textbf{\textcolor{violet}{435}} & \textbf{\textcolor{violet}{745}} & \textbf{\textcolor{violet}{444}} & \textbf{\textcolor{violet}{697}} & \textbf{\textcolor{violet}{547}} \\
    \addlinespace[1.2ex]

    & \textbf{Gemma 8.5B} \\
    & Belief Contradictions     & 121 & 114 & 191 &  103 & 149 & 135 \\
    & Omissions                 & 1 & 0 & 0 & 0 & 0 & 0 \\
    & Unsupported Beliefs       & 62 & 49 & 101 &  53 & 96 & 85 \\
    & False Beliefs             & 2 & 0 & 0 & 1 & 2 & 0 \\
    & \textbf{Total Discrepancies (Gemma)} & \textbf{\textcolor{blue}{186}} & 163 & 292 & 157 & \textbf{\textcolor{blue}{247}}& 220 \\
    \addlinespace[1.2ex]

    & \textbf{Human} \\
    & Belief Contradictions     & 127 & 198 & 181 & 37 & 227 & 188 \\
    & Omissions                 & 52 & 20 & 28 & 52&  128& 103 \\
    & Unsupported Beliefs       & 8 & 4 &  13 &  52 &  88 & 68 \\
    & False Beliefs             & 0 & 0 & 3 & 0 & 0 & 0 \\
    & \textbf{Total Discrepancies (Human)} & 188 & 222 & \textbf{\textcolor{blue}{225}} & \textbf{\textcolor{blue}{141}} & 443 & 359 \\
    \bottomrule
  \end{tabular}
  }
    \caption{LLM Discrepancy analysis results across 6 dialogues for each of 4 discrepancy types
    for all four LLMs. Each row indicates the number of a discrepancy type identified for a given dialogue. An LLM judge compared each model's audio-only mental model trace against the ground truth trace; naive human audio-only traces were evaluated using Claude Sonnet 4 as judge. \textcolor{blue}{Blue} numbers highlight the lowest total number of discrepancies in a dialogue across all models and traces and \textcolor{violet}{purple} numbers highlight the highest total discrepancies in a certain dialogue across all models and traces.}
  \label{table:discrepancy-results}
\end{table}

\subsection{Human Validation of Discrepancy Identification via LLM Judge}
\label{validation}

To examine judge reliability, we manually validated discrepancy
identification on Dialogue 1. For all LLM trace comparisons, the same
model that generated the audio-only trace served as the LLM judge; for
naive human trace comparisons, Claude Sonnet 4 was used as judge.
Table 3 reports overall accuracy per judge on Dialogue 1. 

\begin{table}[H]
\centering
\scriptsize
\begin{tabular}{lccc}
\toprule
\textbf{Judge} & \textbf{Correct} & \textbf{Wrong} & \textbf{Accuracy} \\
\midrule
Claude Sonnet 4 & 383 & 145 & 0.725 \\
Gemma 8.5B & 147 & 39 & 0.791 \\
Naive Human & 98 & 89 & 0.524 \\
o3-mini & 155 & 22 & \textbf{0.876} \\
\bottomrule
\end{tabular}
\caption{Number of correct and incorrect discrepancy identifications for Dialogue 1.}
\label{tab:discrepancy_accuracy_d1}
\end{table}

\section{Results}
\label{results}

We report discrepancy counts across four categories: \textit{belief contradictions}, \textit{omissions}, \textit{unsupported beliefs}, and \textit{false beliefs} for three LLMs and a naive human baseline, each evaluated against ground-truth trace across six situated CReST dialogues.

\begin{figure}[h]
  \centering
  \includegraphics[width=0.85\columnwidth, keepaspectratio]{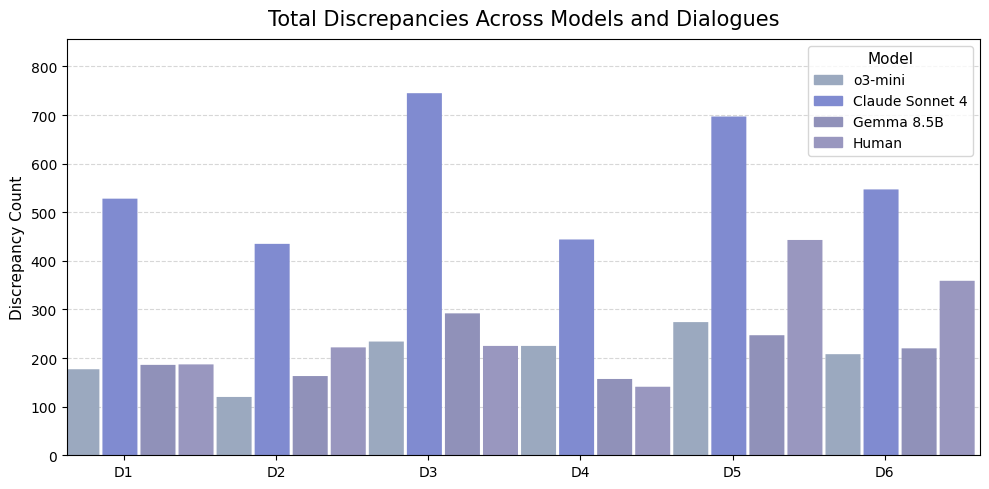}
  \caption{Visualization of the total discrepancies identified between each model's audio-only mental model trace and the ground truth trace. Claude Sonnet 4 produces substantially more discrepancies than other models across all 
dialogues. This pattern admits two interpretations: higher sensitivity to genuine 
discrepancy types (i.e., better recall), or systematic over-generation of unsupported 
inferences (i.e., hallucination). The per-utterance unsupported belief rates in 
Appendix~\ref{discdetails} --- which are consistently the highest for Claude across 
all six dialogues --- suggest the latter is the dominant effect.}
  \label{discrepancy-total}
\end{figure}

\subsection{Discrepancy Trends}

Table~\ref{table:discrepancy-results}, Table \ref{normalized_summary}, and Figure~\ref{discrepancy-total} summarize total discrepancies per model's audio-only trace across all six dialogues. Across both a basic counting metric and normalized weighted score, we see the following.
 \textbf{Claude Sonnet 4} exhibits the highest overall discrepancy counts, surpassing 700 discrepancies in Dialogue 3 and having consistently the lowest discrepancy score across all dialogues as shown in Table \ref{normalized_summary}. The model frequently introduces \textit{unsupported beliefs} and \textit{belief contradictions}, suggesting a tendency toward overgeneralized or speculative reasoning.
\textbf{o3-mini} and \textbf{Gemma 8.5B} produce markedly fewer discrepancies, but with distinct profiles. o3-mini displays high \textit{omission} rates and numerous \textit{unsupported beliefs}, while Gemma identifies substantial \textit{belief contradictions}. False beliefs are notably sparse across all models and traces (Table 6), which we attribute to the open-ended trace generation format: outright factual errors are less likely when audio-only traces record no change rather than commit to an incorrect belief.
\textbf{Naive human annotators} demonstrate a discrepancy distribution that overlaps considerably with LLM profiles, particularly in \textit{belief contradictions} and \textit{omissions}, but outperform LLMs in limiting unsupported beliefs.

Across all models, certain dialogues (D3 and D5) consistently elicit higher discrepancy counts, indicating increased tracking difficulty, likely due to their longer lengths. To validate that discrepancy differences across models are statistically significant, we conducted a two-way ANOVA across 519 model-utterance observations (3 LLM MM trackers x 1,142 utterances, reduced to matched observations across dialogues) (see Appendix C). Results confirm a strong main effect of model (F(2,516)=99.82, p<\$2e-16) with no meaningful effect of utterance position (F(172,346)=0.593, p=0.9999), and pairwise comparisons show Claude differs significantly from both Gemma and o3-mini (both p
<\$2e-16), while Gemma and o3-mini do not differ (p=1.00).

\begin{table}[ht]
  \centering
  \small
  \resizebox{0.85\columnwidth}{!}{
  \begin{tabular}{lcccc}
    \toprule
    Dialogue & o3-mini & Claude Sonnet 4 & Gemma 8.5B & Human \\
    \midrule
    D1 & \textbf{0.917} & 0.104 & 0.896 & 0.894 \\
    D2 & \textbf{1.000} & 0.141 & 0.883 & 0.722 \\
    D3 & 0.910 & 0.000 & 0.807 & \textbf{0.926} \\
    D4 & 0.770 & 0.229 & 0.939 & \textbf{0.978} \\
    D5 & 0.871 & 0.168 & \textbf{0.916} & 0.590 \\
    D6 & \textbf{0.897} & 0.197 & 0.873 & 0.585 \\
    \bottomrule
  \end{tabular}
  }
  \caption{Summary of normalized discrepancy scores \(\mathcal{S}_{m,d}\) for each model across six dialogues with $w_x$ set to 1 for all weights. Bolded values represent the highest for each dialogue. Appendix \ref{discdetails} shows more discrepancy specifics.}
  \label{normalized_summary}
\end{table}

Although Claude Sonnet 4 achieved the lowest accuracy among LLM judges on Dialogue 1, 
it was selected as the judge for naive human traces because it was the only model that 
could evaluate traces it did not itself generate, enabling a consistent cross-condition 
comparison. This introduces a known limitation: Claude's lower precision may inflate 
discrepancy counts for human traces relative to those judged by o3-mini or Gemma. We 
acknowledge this asymmetry and encourage future work to use a single held-out judge 
across all conditions.

\subsection{Comparison to Human Audio-Only Traces}

Naive human audio-only traces contain fewer unsupported beliefs than any
LLM, but comparable or greater numbers of belief contradictions and
omissions. This suggests that, while humans are better at avoiding the
LLMs' hallucinated or unjustified inferences, they are also vulnerable
to erroneous inferences about task state or agent perspective when
lacking access to environmental context. 

In terms of discrepancy counts, the model whose profile most closely
matches human audio-only traces is \textbf{Gemma 8.5B}. However, this
similarity has a trade-off: while Gemma avoids omissions, it more
frequently makes incorrect inferences about ground truth belief
states. We note that per-utterance normalization partially mitigates,
but does not eliminate, differences in trace verbosity across
models; a model that generates fewer field updates by default may
incur fewer omissions by design.

It is worth noting that naive human performance is itself far from perfect: substantial 
belief contradiction and omission rates indicate that the task is genuinely difficult 
without environmental access, not merely an LLM failure mode. This suggests that a 
portion of LLM discrepancy counts reflects irreducible task difficulty rather than 
model-specific deficiency, and future comparisons should interpret LLM performance 
relative to the naive human ceiling rather than against a zero-error baseline.

\subsection{Implications for Situated Theory of Mind}

These findings highlight the nuanced but limited capability of current
LLMs to perform ToM-like inference in grounded, task-based
dialogue. While LLMs can maintain internally consistent belief
structures across dialogue turns, they frequently fail to ground those
beliefs in environmental constraints. The prevalence of unsupported
beliefs and contradictions in Claude suggests limitations in evidence
prioritization and contextual anchoring, while the divergence between
o3-mini's high omission rate and Gemma's near-zero omissions points to
fundamentally different trace generation strategies, underspecification
versus overcommitment, across model families.

While LLMs demonstrate surface-level ToM capabilities in terms of
belief attribution and second-order mental state modeling, their
performance remains qualitatively distinct from that of humans. They
are designed to be driven more by linguistic regularity than grounded
situational reasoning. 

\section{Discussion \& Conclusion}

This study highlights a key limitation of current LLMs: while they can simulate mental 
state attribution, they lack grounded situational reasoning. This claim is supported by 
three converging lines of evidence. First, the qualitative error analysis in 
the Appendix shows that LLM trace errors cluster in utterances 
containing spatial reference or transcription disfluencies, precisely the turns where 
environmental grounding is most needed. Second, per-utterance unsupported belief rates 
(Table~\ref{tab:unsupported_beliefs}) are highest in Dialogues 3 and 5, the longest 
dialogues, suggesting that without environmental anchoring, LLMs accumulate spurious 
inferences over time. Third, false beliefs are rare across all models, indicating that errors arise not 
from confident misidentification but from ungrounded extrapolation. Together, these 
patterns suggest that LLMs infer mental states from linguistic regularity rather than 
from integrated spatial, temporal, and epistemic understanding, producing traces that 
appear internally coherent but fail in settings that require accurate grounding.

\section{Limitations}

Our experiments have several limitations. First, our results depend on the behavior of large language models, which can introduce variability due to non-deterministic outputs, prompt sensitivity, and ongoing model updates. Although we evaluate multiple models from different providers and observe consistent trends across dialogues, some variability in traces and discrepancies may still arise from prompt phrasing and model-specific behavior.

Second, while we include multiple models, our evaluation is not exhaustive. Differences observed across models suggest that discrepancy patterns may vary with model architecture and training data.

Our experimental setup is constrained by the scale of the CReST corpus subset, which includes six dialogues with associated videos. This limited sample size restricts the diversity of interaction patterns and may impact the robustness of our findings. Expanding to larger and more varied dialogue datasets would strengthen the generality of the results.

Additionally, our experimental design does not include a fourth condition in which an LLM generates 
traces with access to both audio and video, as human ground truth annotators do. This 
is a natural extension of the current work and would allow for a more direct comparison 
between human and LLM performance under matched perceptual conditions. We omit this 
condition due to practical constraints at the time of the study, including the cost and 
latency of video-capable multimodal LLM inference at the scale required and variability in 
video-processing support across model providers. We identify this as an important 
direction for future work.

Finally, our framework operates purely on text-based traces without incorporating structured environmental context. In spatially grounded tasks, providing LLM trace generators with explicit spatial representations (e.g., maps or scene layouts) may reduce unsupported beliefs and improve trace accuracy. Exploring such grounded extensions is a promising direction for future work.

\section{Ethical Considerations}
\label{ethical}

In selecting models for this study, we prioritized a balance between
experimental rigor and environmental responsibility. Large-scale
evaluation across dozens of LLMs would have significantly increased
compute time and energy consumption, contributing to the growing
carbon footprint associated with LLM research. Instead, we chose a
representative sample of three high-performing models from major LLM
paradigms: OpenAI’s o3-mini, Claude Sonnet 4, and Gemma. These selections reflect
a diverse cross-section of architectures, providing meaningful comparative insights without
unnecessary ecological impact. 

\section*{Acknowledgements}
This work was in part funded by DARPA contract \#HR001124C0502.

\bibliographystyle{acl_natbib}
\bibliography{custom}

\begin{thebibliography}{40}
\providecommand{\natexlab}[1]{#1}

\bibitem[{Addlesee et~al.(2023)Addlesee, Sieińska, Gunson, Garcia, Dondrup,
  and Lemon}]{Addlesee}
Angus Addlesee, Weronika Sieińska, Nancie Gunson, Daniel~Hernández Garcia,
  Christian Dondrup, and Oliver Lemon. 2023.
\newblock \href {https://doi.org/10.48550/ARXIV.2308.15231} {Multi-party goal
  tracking with llms: Comparing pre-training, fine-tuning, and prompt
  engineering}.

\bibitem[{Andrews et~al.(2023)Andrews, Lilly, Srivastava, and
  Feigh}]{andrews2023role}
Robert~W Andrews, J~Mason Lilly, Divya Srivastava, and Karen~M Feigh. 2023.
\newblock The role of shared mental models in human-ai teams: a theoretical
  review.
\newblock \emph{Theoretical Issues in Ergonomics Science}, 24(2):129--175.

\bibitem[{Bertolazzi et~al.(2024)Bertolazzi, Gatt, and
  Bernardi}]{bertolazzi2024systematicanalysislargelanguage}
Leonardo Bertolazzi, Albert Gatt, and Raffaella Bernardi. 2024.
\newblock \href {https://arxiv.org/abs/2406.11341} {A systematic analysis of
  large language models as soft reasoners: The case of syllogistic inferences}.
\newblock \emph{Preprint}, arXiv:2406.11341.

\bibitem[{Chen et~al.(2024)Chen, Wu, Zhou, Wen, Bi, Jiang, Cao, Hu, Lai, Xiong,
  and Huang}]{chen2024tombenchbenchmarkingtheorymind}
Zhuang Chen, Jincenzi Wu, Jinfeng Zhou, Bosi Wen, Guanqun Bi, Gongyao Jiang,
  Yaru Cao, Mengting Hu, Yunghwei Lai, Zexuan Xiong, and Minlie Huang. 2024.
\newblock \href {https://arxiv.org/abs/2402.15052} {Tombench: Benchmarking
  theory of mind in large language models}.
\newblock \emph{Preprint}, arXiv:2402.15052.

\bibitem[{Eberhard et~al.(2010)Eberhard, Nicholson, K{\"u}bler, Gundersen, and
  Scheutz}]{eberhard}
Kathleen~M Eberhard, Hannele Nicholson, Sandra K{\"u}bler, Susan Gundersen, and
  Matthias Scheutz. 2010.
\newblock The indiana" cooperative remote search task"(crest) corpus.
\newblock In \emph{LREC}.

\bibitem[{Edgar et~al.(2024)Edgar, Aygun, McWilliams, and
  Scheutz}]{edgar2024toward}
Gwendolyn Edgar, Ayca Aygun, Matthew McWilliams, and Matthias Scheutz. 2024.
\newblock Toward genuine robot teammates: Improving human-robot team
  performance beyond shared mental models with proactivity.
\newblock In \emph{Discovering the Frontiers of Human-Robot Interaction:
  Insights and Innovations in Collaboration, Communication, and Control}, pages
  1--22. Springer.

\bibitem[{Gandhi et~al.(2023)Gandhi, Fr{\"a}nken, Gerstenberg, and
  Goodman}]{gandhi2023understanding}
Kanishk Gandhi, Jan-Philipp Fr{\"a}nken, Tobias Gerstenberg, and Noah Goodman.
  2023.
\newblock \href {https://openreview.net/forum?id=8bqjirgxQM} {Understanding
  social reasoning in language models with language models}.
\newblock In \emph{Thirty-seventh Conference on Neural Information Processing
  Systems Datasets and Benchmarks Track}.

\bibitem[{Gervits et~al.(2016)Gervits, Eberhard, and Scheutz}]{gervits2016team}
Felix Gervits, Kathleen Eberhard, and Matthias Scheutz. 2016.
\newblock Team communication as a collaborative process.
\newblock \emph{Frontiers in Robotics and AI}, 3:62.

\bibitem[{Johnson-Laird(1980)}]{JOHNSONLAIRD198071}
P.N. Johnson-Laird. 1980.
\newblock \href {https://doi.org/10.1016/S0364-0213(81)80005-5} {Mental models
  in cognitive science}.
\newblock \emph{Cognitive Science}, 4(1):71--115.

\bibitem[{Jonker et~al.(2010)Jonker, Van~Riemsdijk, and
  Vermeulen}]{jonker2010shared}
Catholijn~M Jonker, M~Birna Van~Riemsdijk, and Bas Vermeulen. 2010.
\newblock Shared mental models: A conceptual analysis.
\newblock In \emph{International workshop on coordination, organizations,
  institutions, and norms in agent systems}, pages 132--151. Springer.

\bibitem[{Kim et~al.(2023)Kim, Sclar, Zhou, Bras, Kim, Choi, and
  Sap}]{Kim_Sclar_Zhou_Bras_Kim_Choi_Sap_2023}
Hyunwoo Kim, Melanie Sclar, Xuhui Zhou, Ronan Bras, Gunhee Kim, Yejin Choi, and
  Maarten Sap. 2023.
\newblock \href {https://doi.org/10.18653/v1/2023.emnlp-main.890} {Fantom: A
  benchmark for stress-testing machine theory of mind in interactions}.
\newblock In \emph{Proceedings of the 2023 Conference on Empirical Methods in
  Natural Language Processing}, page 14397–14413, Singapore. Association for
  Computational Linguistics.

\bibitem[{Kong et~al.(2024)Kong, Zhao, Chen, Li, Qin, Sun, Zhou, Zhou, and
  Sun}]{Kong_Zhao_Chen_Li_Qin_Sun_Zhou_Zhou_Sun_2024}
Aobo Kong, Shiwan Zhao, Hao Chen, Qicheng Li, Yong Qin, Ruiqi Sun, Xin Zhou,
  Jiaming Zhou, and Haoqin Sun. 2024.
\newblock \href {https://doi.org/10.48550/ARXIV.2407.08995} {Self-prompt
  tuning: Enable autonomous role-playing in llms}.

\bibitem[{Kosinski(2024)}]{Kosinski_2024}
Michal Kosinski. 2024.
\newblock \href {https://doi.org/10.1073/pnas.2405460121} {Evaluating large
  language models in theory of mind tasks}.
\newblock \emph{Proceedings of the National Academy of Sciences},
  121(45):e2405460121.
\newblock ArXiv:2302.02083 [cs].

\bibitem[{K{\"u}bler et~al.(2012)K{\"u}bler, Baucom, and
  Scheutz}]{kubler2012parallel}
Sandra K{\"u}bler, Eric Baucom, and Matthias Scheutz. 2012.
\newblock Parallel syntactic annotation in crest.
\newblock \emph{Linguistic Issues in Language Technology}, 7.

\bibitem[{Li et~al.(2026)Li, Gatt, and
  Poesio}]{li2026groundedmisunderstandingsasymmetricdialogue}
Nan Li, Albert Gatt, and Massimo Poesio. 2026.
\newblock \href {https://arxiv.org/abs/2511.03718} {Grounded misunderstandings
  in asymmetric dialogue: A perspectivist annotation scheme for maptask}.
\newblock \emph{Preprint}, arXiv:2511.03718.

\bibitem[{Liu et~al.(2024{\natexlab{a}})Liu, Yao, An, and
  Wang}]{Liu_Yao_An_Wang_2024}
Jiawen Liu, Yuanyuan Yao, Pengcheng An, and Qi~Wang. 2024{\natexlab{a}}.
\newblock \href {https://doi.org/10.1145/3613905.3651008} {Peergpt: Probing the
  roles of llm-based peer agents as team moderators and participants in
  children’s collaborative learning}.
\newblock \emph{Extended Abstracts of the CHI Conference on Human Factors in
  Computing Systems}, page 1–6.

\bibitem[{Liu et~al.(2024{\natexlab{b}})Liu, Zhang, Li, Liu, and
  Yang}]{Liu_Zhang_Li_Liu_Yang_2024}
Zijun Liu, Yanzhe Zhang, Peng Li, Yang Liu, and Diyi Yang. 2024{\natexlab{b}}.
\newblock \href {https://doi.org/10.48550/arXiv.2310.02170} {A dynamic
  llm-powered agent network for task-oriented agent collaboration}.
\newblock (arXiv:2310.02170).
\newblock ArXiv:2310.02170.

\bibitem[{Ma et~al.(2023)Ma, Sansom, Peng, and Chai}]{ma2023towards}
Ziqiao Ma, Jacob Sansom, Run Peng, and Joyce Chai. 2023.
\newblock Towards a holistic landscape of situated theory of mind in large
  language models.
\newblock In \emph{Findings of the Association for Computational Linguistics:
  EMNLP 2023}.

\bibitem[{Manas et~al.(2024)Manas, Zwicklbauer, and Paschke}]{manasetal}
Kumar Manas, Stefan Zwicklbauer, and Adrian Paschke. 2024.
\newblock \href {https://doi.org/10.1109/IV55156.2024.10588650} {Tr2mtl: Llm
  based framework for metric temporal logic formalization of traffic rules}.
\newblock In \emph{2024 IEEE Intelligent Vehicles Symposium (IV)}, pages
  1206--1213.

\bibitem[{Mathieu et~al.(2000)Mathieu, Heffner, Goodwin, Salas, and
  Cannon-Bowers}]{mathieu2000influence}
John~E Mathieu, Tonia~S Heffner, Gerald~F Goodwin, Eduardo Salas, and Janis~A
  Cannon-Bowers. 2000.
\newblock The influence of shared mental models on team process and
  performance.
\newblock \emph{Journal of applied psychology}, 85(2):273.

\bibitem[{Mondorf and
  Plank(2024)}]{mondorf2024accuracyevaluatingreasoningbehavior}
Philipp Mondorf and Barbara Plank. 2024.
\newblock \href {https://arxiv.org/abs/2404.01869} {Beyond accuracy: Evaluating
  the reasoning behavior of large language models -- a survey}.
\newblock \emph{Preprint}, arXiv:2404.01869.

\bibitem[{Paige et~al.(2024)Paige, Soubki, Murzaku, Rambow, and
  Brennan}]{Paige_Soubki_Murzaku_Rambow_Brennan_2024}
Amie Paige, Adil Soubki, John Murzaku, Owen Rambow, and Susan~E. Brennan. 2024.
\newblock \href {https://doi.org/10.18653/v1/2024.sigdial-1.18} {Training llms
  to recognize hedges in dialogues about roadrunner cartoons}.
\newblock In \emph{Proceedings of the 25th Annual Meeting of the Special
  Interest Group on Discourse and Dialogue}, page 204–215, Kyoto, Japan.
  Association for Computational Linguistics.

\bibitem[{Palmer et~al.()Palmer, Zhu, Lai, VanderHoeven, Bradford, Khebour,
  Mabrey, Fitzgerald, Krishnaswamy, Palmer et~al.}]{palmerspeech}
D~Palmer, Y~Zhu, K~Lai, H~VanderHoeven, M~Bradford, I~Khebour, C~Mabrey,
  J~Fitzgerald, N~Krishnaswamy, M~Palmer, et~al.
\newblock Speech is not enough: interpreting nonverbal indicators of common
  knowledge and engagement (2024).

\bibitem[{Pan et~al.(2023)Pan, Albalak, Wang, and Wang}]{pan2023logic}
Liangming Pan, Alon Albalak, Xinyi Wang, and William~Yang Wang. 2023.
\newblock Logic-lm: Empowering large language models with symbolic solvers for
  faithful logical reasoning.
\newblock \emph{arXiv preprint arXiv:2305.12295}.

\bibitem[{Pangakis and Wolken(2024)}]{Pangakis_Wolken_2024}
Nicholas Pangakis and Samuel Wolken. 2024.
\newblock \href {https://doi.org/10.48550/ARXIV.2406.17633} {Knowledge
  distillation in automated annotation: Supervised text classification with
  llm-generated training labels}.

\bibitem[{Sarkar et~al.(2025)Sarkar, Srikanth, Hudson, Rudinger, Bonial, and
  Resnik}]{sarkar2025understandingcommongroundmisalignment}
Rupak Sarkar, Neha Srikanth, Taylor Hudson, Rachel Rudinger, Claire Bonial, and
  Philip Resnik. 2025.
\newblock \href {https://arxiv.org/abs/2503.12370} {Understanding common ground
  misalignment in goal-oriented dialog: A case-study with ubuntu chat logs}.
\newblock \emph{Preprint}, arXiv:2503.12370.

\bibitem[{Sarıtaş et~al.(2025)Sarıtaş, Tezören, and
  Durmazkeser}]{sarıtas2025systematicreviewevaluationlarge}
Karahan Sarıtaş, Kıvanç Tezören, and Yavuz Durmazkeser. 2025.
\newblock \href {https://arxiv.org/abs/2502.08796} {A systematic review on the
  evaluation of large language models in theory of mind tasks}.
\newblock \emph{Preprint}, arXiv:2502.08796.

\bibitem[{Schelble et~al.(2022)Schelble, Flathmann, McNeese, Freeman, and
  Mallick}]{schelble2022let}
Beau~G Schelble, Christopher Flathmann, Nathan~J McNeese, Guo Freeman, and
  Rohit Mallick. 2022.
\newblock Let's think together! assessing shared mental models, performance,
  and trust in human-agent teams.
\newblock \emph{Proceedings of the ACM on Human-Computer Interaction},
  6(GROUP):1--29.

\bibitem[{Scheutz et~al.(2017)Scheutz, DeLoach, and
  Adams}]{scheutz2017framework}
Matthias Scheutz, Scott~A DeLoach, and Julie~A Adams. 2017.
\newblock A framework for developing and using shared mental models in
  human-agent teams.
\newblock \emph{Journal of Cognitive Engineering and Decision Making},
  11(3):203--224.

\bibitem[{Scheutz et~al.(2024)Scheutz, Oosterveld, Peterson, Wyss, and
  Krause}]{scheutz2024multi}
Matthias Scheutz, Bradley Oosterveld, John Peterson, Eric Wyss, and Evan
  Krause. 2024.
\newblock A multi-robot architecture framework for effective robot teammates in
  mixed-initiative teams.
\newblock In \emph{Proceedings of the 2024 International Symposium on
  Technological Advances in Human-Robot Interaction}, pages 74--82.

\bibitem[{Shojaee*† et~al.(2025)Shojaee*†, Mirzadeh*, Alizadeh, Horton,
  Bengio, and Farajtabar}]{illusion-of-thinking}
Parshin Shojaee*†, Iman Mirzadeh*, Keivan Alizadeh, Maxwell Horton, Samy
  Bengio, and Mehrdad Farajtabar. 2025.
\newblock \href
  {https://ml-site.cdn-apple.com/papers/the-illusion-of-thinking.pdf} {The
  illusion of thinking: Understanding the strengths and limitations of
  reasoning models via the lens of problem complexity}.

\bibitem[{Stechly et~al.(2024)Stechly, Valmeekam, and
  Kambhampati}]{stechly2024selfverificationlimitationslargelanguage}
Kaya Stechly, Karthik Valmeekam, and Subbarao Kambhampati. 2024.
\newblock \href {https://arxiv.org/abs/2402.08115} {On the self-verification
  limitations of large language models on reasoning and planning tasks}.
\newblock \emph{Preprint}, arXiv:2402.08115.

\bibitem[{Strachan et~al.(2024)Strachan, Albergo, Borghini, Pansardi, Scaliti,
  Gupta, Saxena, Rufo, Panzeri, Manzi et~al.}]{strachan2024testing}
James~WA Strachan, Dalila Albergo, Giulia Borghini, Oriana Pansardi, Eugenio
  Scaliti, Saurabh Gupta, Krati Saxena, Alessandro Rufo, Stefano Panzeri, Guido
  Manzi, et~al. 2024.
\newblock Testing theory of mind in large language models and humans.
\newblock \emph{Nature Human Behaviour}, 8(7):1285--1295.

\bibitem[{Ullman(2023)}]{Ullman_2023}
Tomer Ullman. 2023.
\newblock \href {https://doi.org/10.48550/arXiv.2302.08399} {Large language
  models fail on trivial alterations to theory-of-mind tasks}.
\newblock (arXiv:2302.08399).
\newblock ArXiv:2302.08399 [cs].

\bibitem[{Weissweiler et~al.(2024)Weissweiler, Köksal, and
  Schütze}]{Weissweiler_Koksal_Schutze_2024}
Leonie Weissweiler, Abdullatif Köksal, and Hinrich Schütze. 2024.
\newblock \href {https://doi.org/10.48550/ARXIV.2403.06965} {Hybrid human-llm
  corpus construction and llm evaluation for rare linguistic phenomena}.

\bibitem[{Wu et~al.(2024)Wu, Mu, Zhou, Ma, Chen, and
  Liu}]{Wu_Mu_Zhou_Ma_Chen_Liu_2024}
Pengying Wu, Yao Mu, Kangjie Zhou, Ji~Ma, Junting Chen, and Chang Liu. 2024.
\newblock \href {https://doi.org/10.48550/ARXIV.2407.00632} {Camon: Cooperative
  agents for multi-object navigation with llm-based conversations}.

\bibitem[{Yu et~al.(2024)Yu, Li, Su, and Fuoli}]{Yu_Li_Su_Fuoli_2024}
Danni Yu, Luyang Li, Hang Su, and Matteo Fuoli. 2024.
\newblock \href {https://doi.org/10.48550/arXiv.2305.08339} {Assessing the
  potential of ai-assisted pragmatic annotation: The case of apologies}.
\newblock (arXiv:2305.08339).
\newblock ArXiv:2305.08339.

\bibitem[{Zhao et~al.(2024)Zhao, Thai, Zhang, Hu, Ba, Zhou, Cai, Liu, and
  Sun}]{zhao2024decoratelm}
Ranchi Zhao, Zhen~Leng Thai, Yifan Zhang, Shengding Hu, Yunqi Ba, Jie Zhou, Jie
  Cai, Zhiyuan Liu, and Maosong Sun. 2024.
\newblock Decoratelm: Data engineering through corpus rating, tagging, and
  editing with language models.
\newblock \emph{arXiv preprint arXiv:2410.05639}.

\bibitem[{Zhou et~al.(2024)Zhou, Staats, Li, Szegedy, Weinberger, and
  Wu}]{zhou2024don}
Jin~Peng Zhou, Charles Staats, Wenda Li, Christian Szegedy, Kilian~Q
  Weinberger, and Yuhuai Wu. 2024.
\newblock Don't trust: Verify--grounding llm quantitative reasoning with
  autoformalization.
\newblock \emph{arXiv preprint arXiv:2403.18120}.

\bibitem[{Zhu et~al.(2025)Zhu, Lai, Khebour, Verhagen, Krishnaswamy, and
  Pustejovsky}]{zhu2025multimodal}
Yifan Zhu, Kenneth Lai, Ibrahim Khebour, Marc Verhagen, Nikhil Krishnaswamy,
  and James Pustejovsky. 2025.
\newblock Multimodal situational awareness: Neuro-symbolic ai for real-time hci
  in the classroom.
\newblock In \emph{International Conference on Human-Computer Interaction},
  pages 454--464. Springer.

\end{thebibliography}

\appendix

\clearpage           
\onecolumn
\section{Full Prompts}
\label{sec:full_prompts}

The following section outlines our prompts used for the experiments. These prompts were all given the exact same way to all models to compare models as effectively as possible. To note, we tried many different prompts during this project including persona prompting, chain of thought prompting, and more. Across initial experiments we determined that these prompts performed the best and therefore we ran primary experiments with them. Note: The prompt below uses the term 'annotator belief' as a shorthand label within the JSON output schema. In our framework, this corresponds to the belief recorded in the audio-only mental model trace, as described in Section 4.

\begin{tcolorbox}[
  colback=pastelblue,
  colframe=darkblue,
  title=Annotation Prompt,
  fonttitle=\bfseries,
  width=\textwidth,
  arc=3mm,
  boxrule=0.4pt,
  coltitle=black,
  enhanced,
  left=2pt, right=2pt, top=2pt, bottom=2pt, breakable
]

\begin{scriptsize}
\begin{verbatim}
You are an annotator of a dialogue of a 2-person team working together on a search mission.
You will be provided with several rules and examples of how to perform belief, commitment, and goal updates based on dialogue history. 
You are only tracking the beliefs, goals, and commitments which share information related to the searcher's location and level of
task completion.

TEAM MEMBERS:
1. Searcher: The searcher is in a remotely located real world environment of a floor in a building with many rooms, boxes, and doors.
2. Director: The director is local but has a static (but somewhat flawed) map of the floor showing some of the boxes, rooms and doors, 
and is communicating and coordinating via audio with the searcher.

YOUR TASK:
You will carefully listen to the conversation (current dialogue move and recent dialogue history) and logically deduce/induce/abduce 
beliefs, goals, and commitments held by the director and searcher suggested by the dialogue move in context of the dialogue history. 
You will be given a prior state of the world and your task is to decide what beliefs, goals, and commitments to add and what to remove 
to this state based on what was said in the dialogue move and history. 
You have to deduce the beliefs, goals, and commitments yourself. You should be thinking from the perspective of the searcher when 
determining their beliefs, goals, and commitments and vice versa for the director. 
This does not mean that the searcher and director beliefs will align. 

For each utterance, you must return a JSON object with the following structure:
{
    "speaker": <speaker>,
    "utterance": <utterance>,
    "start": <start>,
    "end": <end>,
    "Annotation": {
        "Searcher believes": <string>,
        "Director believes": <string>,
        "2nd order: Searcher believes that the director believes": <string>,
        "2nd order: Director believes that the searcher believes": <string>,
        "Searcher has committed to": <string>,
        "Director has committed to": <string>,
        "Director's goal is": <string>,
        "Searcher's goal is": <string>,
        "Common Belief": <string>
    }
}
For any field where you do not think there is an update, write "no change".
For the "Common Belief" field, if there is a clear belief that both agents share, summarize it in 1-2 sentences. Otherwise, write 
"No change".

WHAT ARE BELIEFS, GOALS, AND COMMITMENTS?
Beliefs: the understanding and assumptions that an agent holds about a situation, task, or team dynamic I.E. where an agent is, 
what they're doing, the state of the world
Goals: the objectives or desired outcomes that an agent strives to achieve
Commitment: agreement of an agent to their individual responsibilities, actions, and expected contributions towards achieving 
a goal.

EXAMPLES OF HOW TO RESPOND:

EXAMPLE 1:

DIALOGUE HISTORY: 
Searcher: "So we're supposed to get the green boxes?"
Director: "I think so."

CURRENT DIALOGUE MOVE:
Searcher: "Okay."

OUTPUT: 
{
    "speaker": "Searcher",
    "utterance": "Okay.",
    "start": <start>,
    "end": <end>,
    "Annotation": {
        "Searcher believes": "no change",
        "Director believes": "no change",
        "2nd order: Searcher believes that the director believes": "no change",
        "2nd order: Director believes that the searcher believes": "no change",
        "Searcher has committed to": "no change",
        "Director has committed to": "no change",
        "Director's goal is": "no change",
        "Searcher's goal is": "The searcher's goal is to get the green boxes.",
        "Common Belief": "Both the searcher and director have the shared goal to get green boxes."
    }
}

EXAMPLE 2:

DIALOGUE HISTORY: 
Director: "okay . in front of your there should be a platform with steps going  up?"
Searcher: "right"
Director: "okay so make a right turn"
Searcher: "kay"
Director: "okay a:nd walk into the next room . through the- there should be an open door there"
Searcher: "right"
Director: "okay walk into the next room"
Searcher: "kay"
Director: "okay . now you should be in a roo:m wi:th . like on the right there should be like two cubicles"
Searcher: "yea there's two like two little rooms"
Director: "ah-"
Director: "okay and straight in front of you should be: . filing cabinet?"

CURRENT DIALOGUE MOVE:
Searcher: "yes"

OUTPUT:
{
    "speaker": "Searcher",
    "utterance": "yes",
    "start": <start>,
    "end": <end>,
    "Annotation": {
        "Searcher believes": "The searcher believes that there is a filing cabinet in front of them.",
        "Director believes": "no change",
        "2nd order: Searcher believes that the director believes": "The searcher believes that the director believes there is a 
        filing cabinet in front of them.",
        "2nd order: Director believes that the searcher believes": "The director believes the searcher believes there is a filing 
        cabinet in front of the searcher.",
        "Searcher has committed to": "no change",
        "Director has committed to": "no change",
        "Director's goal is": "no change",
        "Searcher's goal is": "no change",
        "Common Belief": "Both agents believe there is a filing cabinet in front of the searcher."
    }
}

Learn from these examples and generalize to new cases as you reason through the dialogue you are given. 
You need to infer new states from the past ones. Think from the perspectives of both the director and searcher and what 
they're thinking at each time step. 

The point of these annotations is to maintain an understanding of the shared mental models of the searcher and director. 
We want to track what the searcher and director each independently think is going on in the task based on their perception 
of the searcher's location.
This allows us to investigate the coherence of the team. 

You should only use the following verbs when trying to identify a belief, goal, or commitment:
    at,
    in,
    holding,
    connects, 
    near,
    right of,
    in front of, 
    on, 
    across from,
    get,
    go, 
    turn,
    find

TAKE THESE STEPS TO PERFORM YOUR TASK ON EACH MOVE:
1. Identify the speaker of the current dialogue move.
2. Identify the dialogue act of the move. Could be one of question | assertion/statement | command/request | offer | promise | 
acknowledgement | greeting. This can influence what you conclude about the beliefs. For example, questions about p suggest 
that the speaker did not previously believe p.  
3. Identify the current state of the world.
4. Based on the current dialogue move and the history, describe what beliefs, goals, and commitments of the director and 
searcher you can infer (implicitly or explicitly). Use ONLY the phrasing "The [searcher/director] believes", "The [searcher/director] 
is committed to", "The [searcher/director]'s goal is" to begin the annotation.
5. Reason through this and rationalize why you included the updates that you chose to include.

If no updates are needed, write "no change" for every field in the Annotation object. 

Output MUST be a JSON string, and NOT markdown.
IMPORTANT: Output ONLY the JSON object, with no explanation or commentary. Do not include any reasoning or 
markdown formatting.
\end{verbatim}
\end{scriptsize}

\end{tcolorbox}

\begin{tcolorbox}[
  colback=pastelblue,
  colframe=darkblue,
  title=Discrepancy Detection Prompt,
  fonttitle=\bfseries,
  width=\textwidth,
  arc=3mm,
  boxrule=0.4pt,
  coltitle=black,
  enhanced,
  left=2pt, right=2pt, top=2pt, bottom=2pt,
  breakable,
]

\begin{scriptsize}
\begin{verbatim}
You are given two sets of annotations representing the Shared Mental Models (SMMs) of two agents in a dialogue. Your task is to 
identify and classify discrepancies between the two agents' mental models using the following five types:

1. Belief Contradiction: One annotator identifies a belief b, and the other identifies not b.
2. False Belief: An annotator believes something that contradicts the known ground truth.
3. Omission: The ground truth includes a belief that the annotator simply omits.
4. Unsupported Belief: A belief not verifiable from the context or ground truth; lacks support either way.

For each discrepancy:
- Classify it into one or more of the types above.
- Clearly state what each agent believes, using plain English.
- Provide a short explanation.

Output Format:
You must output a single JSON object with the following structure. For each discrepancy found, create an object with the keys: 
'Discrepancy Type', 'Ground Truth Belief', 'Annotator Belief', and 'Explanation'. If no discrepancies are found, return an empty array.

Required JSON structure:
    "Discrepancies": [
        {
            "Discrepancy Type": "Belief Contradiction",
            "Ground Truth Belief": "belief from ground truth",
            "Annotator Belief": "belief from annotator", 
            "Explanation": "explanation of the discrepancy"
        }
    ]

Examples:

Example 1: Belief Contradiction
{
    "Discrepancies": [
        {
            "Discrepancy Type": "Belief Contradiction",
            "Ground Truth Belief": "The searcher is at the cardboard box",
            "Annotator Belief": "The searcher is at room 1",
            "Explanation": "The ground truth indicates the searcher is at the cardboard box, while the annotator believes the searcher 
            is at room 1. These beliefs are directly contradictory."
        }
    ]
}

Example 2: False Belief
{
    "Discrepancies": [
        {
            "Discrepancy Type": "False Belief",
            "Ground Truth Belief": "The searcher is in room r",
            "Annotator Belief": "The searcher is not in room r",
            "Explanation": "The annotator believes the searcher is not in room r, but the ground truth indicates that the searcher is 
            actually there."
        }
    ]
}

Example 3: Omission
{
    "Discrepancies": [
        {
            "Discrepancy Type": "Omission",
            "Ground Truth Belief": "The director's goal is for the searcher to go to corridor 2",
            "Annotator Belief": "No mention of director's goal",
            "Explanation": "The ground truth includes a goal for the searcher to go to corridor 2, while the annotator omits this 
            information entirely."
        }
    ]
}

Example 4: Unsupported Belief
{
    "Discrepancies": [
        {
            "Discrepancy Type": "Unsupported Belief",
            "Ground Truth Belief": "No specific belief about other agent's thoughts",
            "Annotator Belief": "The annotator believes that the other agent thinks the searcher is in room 1",
            "Explanation": "The annotator makes an assumption about the other agent's belief that cannot be verified from the given 
            dialogue or any known facts."
        }
    ]
}

Example 5: No Discrepancies
{
    "Discrepancies": []
}
\end{verbatim}
\end{scriptsize}

\end{tcolorbox}

\definecolor{pastelblue}{RGB}{230,243,255}

\section{CReST Dataset Information}
\label{sec:extra}

\begin{tcolorbox}[
  colback=pastelblue,
  colframe=darkblue,
  coltitle=black,
  title=Map of Environment,
  fonttitle=\bfseries,
  width=\columnwidth,
  arc=3mm,
  boxrule=0.4pt,
  enhanced,
  left=4pt, right=4pt, top=4pt, bottom=4pt,
  breakable,
]

\vspace{1mm}
\includegraphics[width=\columnwidth]{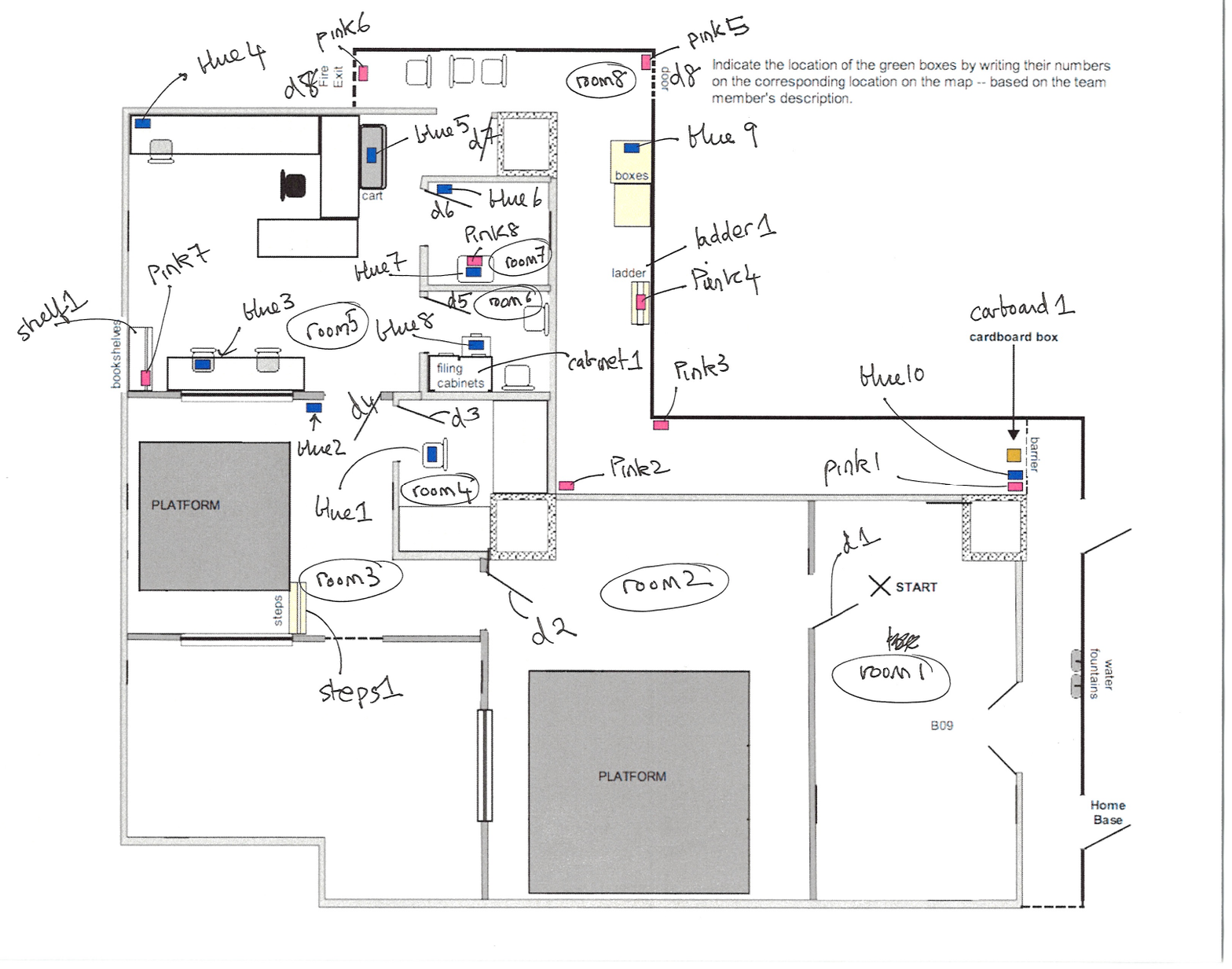}
\vspace{1mm}

\noindent
The map supplied to the director in the CReST experiments, which is flawed in some minor ways so as to induce confusion between the searcher and director. The director guides the searcher through this environment to the blue boxes as the searcher scans for green boxes, and this map's locations of blue boxes is partially incorrect. Here, we labeled the locations of objects and rooms as given to the LLM, though that was not given to the director at the start of the experiment. The searcher is in this environment and is not given this information.

\end{tcolorbox}

\vspace{1em}

\begin{tcolorbox}[
  colback=pastelblue,
  colframe=darkblue,
  coltitle=black,
  title=Video Screenshot of Environment,
  fonttitle=\bfseries,
  width=\columnwidth,
  arc=3mm,
  boxrule=0.4pt,
  enhanced,
  left=4pt, right=4pt, top=4pt, bottom=4pt,
  breakable,
]

\vspace{1mm}
\includegraphics[width=\columnwidth]{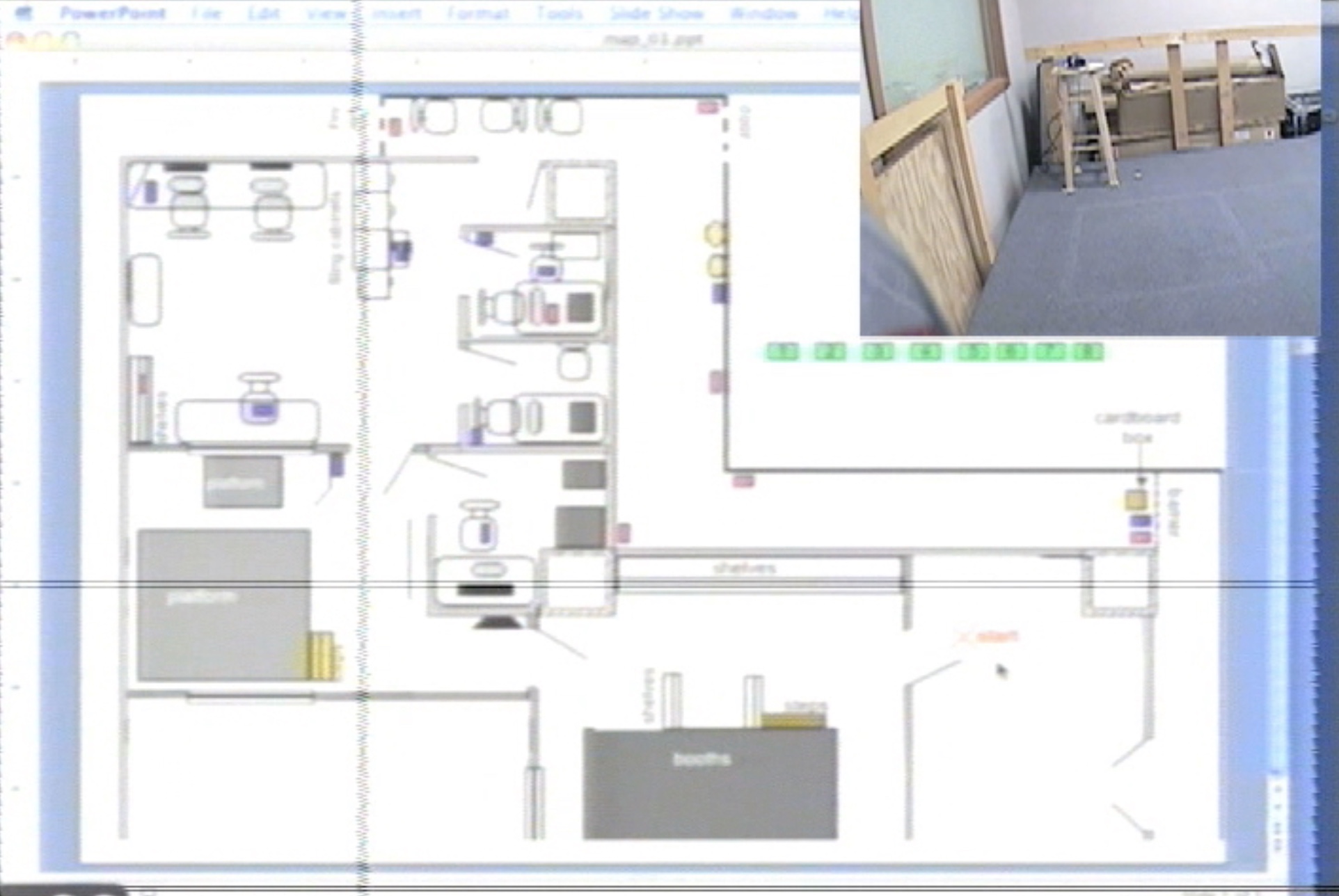}
\vspace{1mm}

\noindent
Here we have supplied a screenshot from a video of the CReST corpus. Videos such as this one were used to create the dataset of ground truth mental model traces for each dialogue, which was corroborated by two human annotators. Here, one can see that there is an image of the map in the director's view and the black lines forming a cross show the searcher's current location, which dynamically moves during the video to update the real-time searcher's location. In the top right hand corner is the headcam video from the searcher's perspective, which shows exactly what they are looking at. Because of these videos, we were able to establish a ground truth understanding of the environment external to the naive human and LLM trace generators, which we use in our evaluations of the LLM discrepancy finder.

\end{tcolorbox}

\twocolumn

\section{Statistical Validation of Results}

To validate the statistical significance of this study, we ran ANOVA tests between the two conditions of naive human and LLM audio-only traces, as evaluated by the LLM judge. Our two-way ANOVA is pasted below, with the p-value of less than 2e-16.

\begin{figure}[h]
    \centering
    \includegraphics[width=\columnwidth]{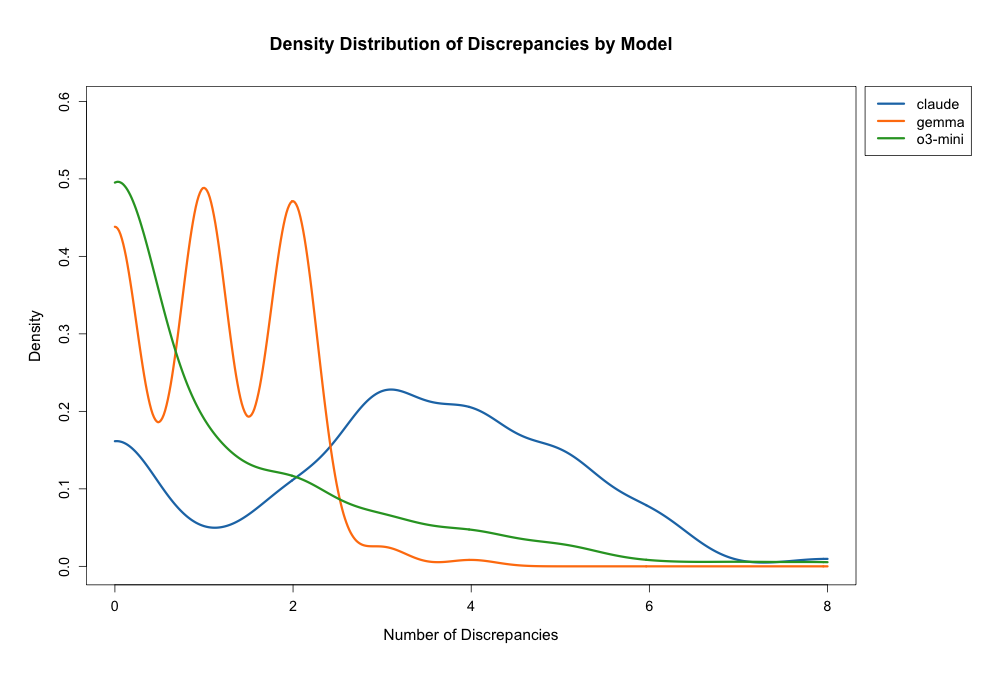}
    \caption{Violin plot of discrepancy density across LLMs-As-A-Judge.}
    \label{fig:violin}
\end{figure}

Most notably, we found a model effect where LLM judges showed a difference in identified discrepancies. Across 519 model-utterance observations, the violin plot shows a clear separation in discrepancy distributions by model: Claude is centered at substantially higher discrepancy counts and has a broader spread, whereas gemma and o3-mini are concentrated at low counts with strong overlap (both near 1 discrepancy on average). This visual pattern is confirmed by ANOVA, which found a strong main effect of model on discrepancies (F(2,516)=99.82, p<2e-16), but no meaningful effect of utterance number (F(172,346)=0.593, p=0.9999). Pairwise comparisons (Bonferroni-adjusted) further indicate that Claude differs significantly from both Gemma and o3-mini (both p<2e-16), while gemma and o3-mini do not differ (p=1.00), supporting the conclusion that discrepancy burden is primarily model-driven rather than position-in-dialogue-driven.

\section{Discrepancy Visualizations}
\label{discrepancies}

\centering
\includegraphics[width=\columnwidth]{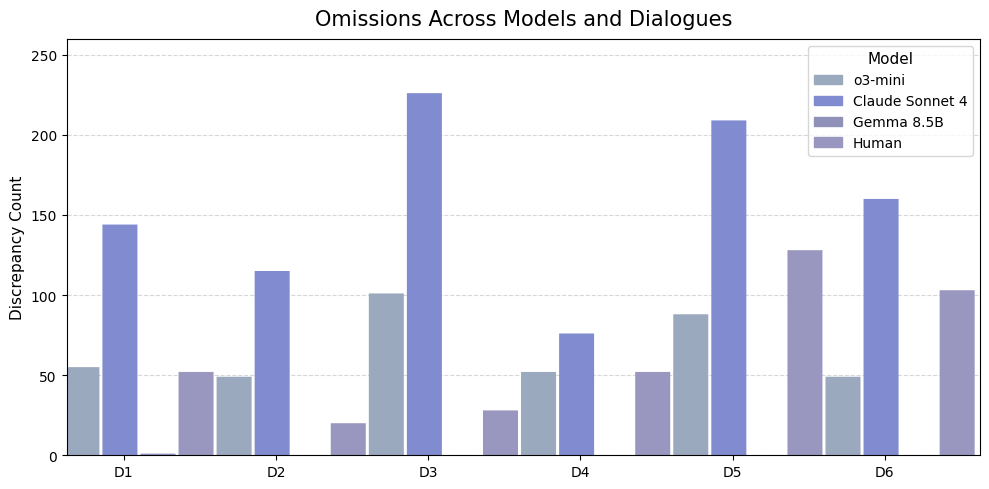}
\vspace{2mm}

\noindent
\textit{This figure shows the number of omissions identified by each LLM across the dialogue corpus.}

\vspace{1em}

\centering
\includegraphics[width=\columnwidth]{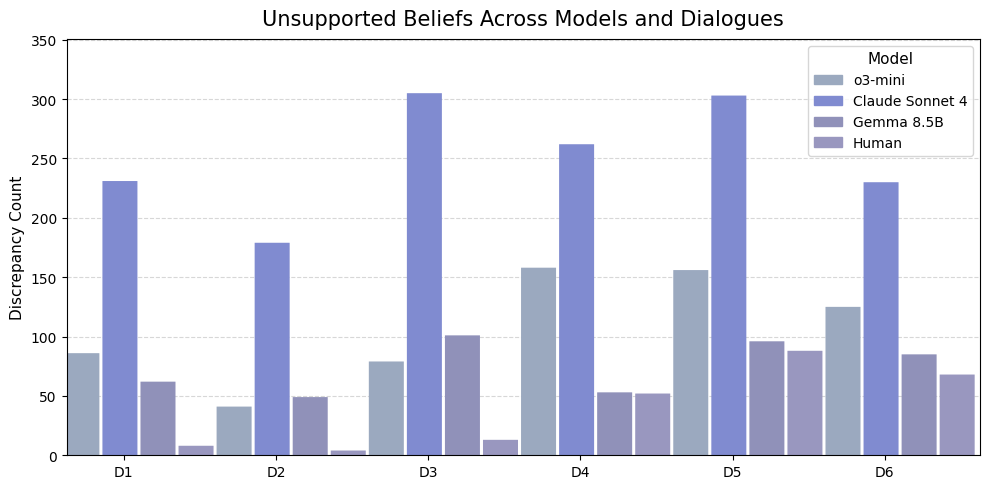}
\vspace{2mm}

\noindent
\textit{This figure visualizes the number of unsupported beliefs returned by each LLM.}

\vspace{1em}

\centering
\includegraphics[width=\columnwidth]{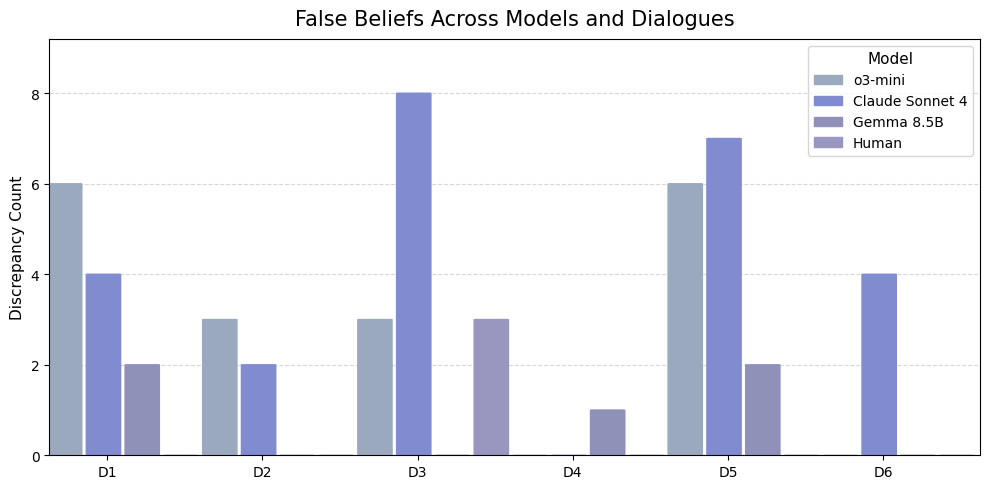}
\vspace{2mm}

\noindent
\textit{This figure displays false belief discrepancies identified when comparing LLM audio-only traces against the ground truth trace.}

\vspace{1em}

\includegraphics[width=\columnwidth]{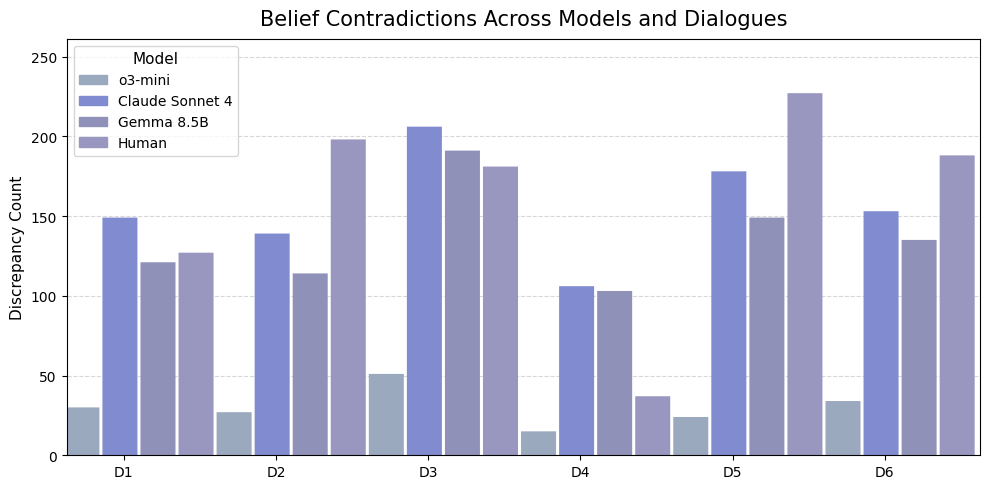}
\vspace{2mm}

\noindent
\textit{This figure shows contradictions between ground truth and LLM-derived beliefs.}

\section{Discrepancy Details}
\label{discdetails}
\par
\raggedright{These tables display, for each discrepancy type, the raw count of errors divided by the number of utterances in each dialogue, yielding a fair measure of how often that error occurs per turn. Presenting each discrepancy type in its own table makes it easy to spot patterns in belief contradictions, false beliefs, unsupported beliefs, and omissions before any further scaling.}

\vspace{2mm}
\resizebox{\columnwidth}{!}{
\begin{tabular}{lcccc}
  \toprule
  Dialogue & o3-mini & Claude Sonnet 4 & Gemma 8.5B & Naive Human \\
  \midrule
  D1 & 0.173 & 0.861 & 0.699 & 0.734 \\
  D2 & 0.184 & 0.946 & 0.776 & 1.347 \\
  D3 & 0.227 & 0.916 & 0.849 & 0.804 \\
  D4 & 0.093 & 0.654 & 0.636 & 0.228 \\
  D5 & 0.100 & 0.739 & 0.618 & 0.942 \\
  D6 & 0.175 & 0.789 & 0.696 & 0.969 \\
  \bottomrule
\end{tabular}
}
\captionof{table}{Per-utterance rates for Belief Contradictions}
\label{tab:belief_contradictions}

\vspace{4mm}
\resizebox{\columnwidth}{!}{
\begin{tabular}{lcccc}
  \toprule
  Dialogue & o3-mini & Claude Sonnet 4 & Gemma 8.5B & Naive Human \\
  \midrule
  D1 & 0.035 & 0.023 & 0.012 & 0.000 \\
  D2 & 0.020 & 0.014 & 0.000 & 0.000 \\
  D3 & 0.013 & 0.036 & 0.000 & 0.013 \\
  D4 & 0.000 & 0.000 & 0.006 & 0.000 \\
  D5 & 0.025 & 0.029 & 0.008 & 0.000 \\
  D6 & 0.000 & 0.021 & 0.000 & 0.000 \\
  \bottomrule
\end{tabular}
}
\captionof{table}{Per-utterance rates for False Beliefs}
\label{tab:false_beliefs}

\vspace{4mm}

\resizebox{\columnwidth}{!}{
\begin{tabular}{lcccc}
  \toprule
  Dialogue & o3-mini & Claude Sonnet 4 & Gemma 8.5B & Naive Human \\
  \midrule
  D1 & 0.497 & 1.335 & 0.358 & 0.046 \\
  D2 & 0.279 & 1.218 & 0.333 & 0.027 \\
  D3 & 0.351 & 1.356 & 0.449 & 0.058 \\
  D4 & 0.975 & 1.617 & 0.327 & 0.321 \\
  D5 & 0.647 & 1.257 & 0.398 & 0.365 \\
  D6 & 0.644 & 1.186 & 0.438 & 0.351 \\
  \bottomrule
\end{tabular}
}
\captionof{table}{Per-utterance rates for Unsupported Beliefs}
\label{tab:unsupported_beliefs}

\vspace{4mm}
\resizebox{\columnwidth}{!}{
\begin{tabular}{lcccc}
  \toprule
  Dialogue & o3-mini & Claude Sonnet 4 & Gemma 8.5B & Naive Human \\
  \midrule
  D1 & 0.318 & 0.832 & 0.006 & 0.301 \\
  D2 & 0.333 & 0.782 & 0.000 & 0.136 \\
  D3 & 0.449 & 1.004 & 0.000 & 0.124 \\
  D4 & 0.321 & 0.469 & 0.000 & 0.321 \\
  D5 & 0.365 & 0.867 & 0.000 & 0.531 \\
  D6 & 0.253 & 0.825 & 0.000 & 0.531 \\
  \bottomrule
\end{tabular}
}
\captionof{table}{Per-utterance rates for Omissions}
\label{tab:omissions}

These tables show an example of our metric applied to our dataset. They show the normalized scores for each model across a dialogue, taking into account the per-utterance scores above. This condenses all four discrepancy types into a single metric $\mathcal{S}_{m,d}$ for each model and dialogue, linearly scaled to the range [0, 1] so that higher values indicate cleaner performance.  By dividing the weighted raw discrepancy counts by dialogue length and then normalizing against the global minimum and maximum scores, these tables provide a directly comparable measure of overall error rate per utterance.  Viewing one table per model highlights which dialogues each system handles most accurately. This metric may be used with assigned weights, but for our use case they were all set to 1. 

\vspace{2mm}
\centering
\begin{tabular}{lc}
  \toprule
  Dialogue & Normalized Score \\
  \midrule
  D1 & 0.917 \\
  D2 & 1.000 \\
  D3 & 0.910 \\
  D4 & 0.770 \\
  D5 & 0.871 \\
  D6 & 0.897 \\
  \bottomrule
\end{tabular}
\captionof{table}{Normalized scores for \texttt{o3-mini}}
\label{tab:norm_o3mini}

\vspace{4mm}
\begin{tabular}{lc}
  \toprule
  Dialogue & Normalized Score \\
  \midrule
  D1 & 0.104 \\
  D2 & 0.141 \\
  D3 & 0.000 \\
  D4 & 0.229 \\
  D5 & 0.168 \\
  D6 & 0.197 \\
  \bottomrule
\end{tabular}

\captionof{table}{Normalized scores for \texttt{Claude Sonnet 4}}
\label{tab:norm_claude}

\begin{tabular}{lc}
  \toprule
  Dialogue & Normalized Score \\
  \midrule
  D1 & 0.896 \\
  D2 & 0.883 \\
  D3 & 0.807 \\
  D4 & 0.939 \\
  D5 & 0.916 \\
  D6 & 0.873 \\
  \bottomrule
\end{tabular}

\captionof{table}{Normalized scores for \texttt{Gemma 8.5B}}
\label{tab:norm_gemma}

\vspace{2mm}
\begin{tabular}{lc}
  \toprule
  Dialogue & Normalized Score \\
  \midrule
  D1 & 0.894 \\
  D2 & 0.722 \\
  D3 & 0.926 \\
  D4 & 0.978 \\
  D5 & 0.590 \\
  D6 & 0.585 \\
  \bottomrule
\end{tabular}

\captionof{table}{Normalized scores for \texttt{Naive Human}}
\label{tab:norm_human}

\onecolumn
\raggedright
\section{Qualitative Error Analysis}

To substantiate our claims about spatial reasoning and disfluency failures, we present representative examples drawn from the human-validated discrepancy set for Dialogue 1. For each example we show the utterance, the ground truth trace, the LLM trace, the discrepancy type identified by the judge, and a brief explanation. Examples are selected to illustrate the three most common failure modes: spatial misgrounding, disfluency-induced hallucination, and correct trace for contrast.

\begin{table*}[h]
\small
\centering
\renewcommand{\arraystretch}{1.4}
\begin{tabular}{p{2.8cm} p{1.5cm} p{2.2cm} p{4.8cm}}
\hline
\textbf{Utterance} & \textbf{Model} & \textbf{Discrepancy Type} & 
\textbf{Evidence} \\
\hline
\textit{``yes''} (t=27.18s) & Gemma 8.5B & Unsupported Belief & 
The audio-only trace introduces a spatial second-order belief (that the searcher knows the platform and cubicle location) that is absent from the ground truth trace. \\
\hline
\textit{``and uh there's a cubicle on your right hand side and 
then straight in front of you there's also a door?''} (t=22.45s) & 
o3-mini & Unsupported Belief & 
The audio-only trace introduces a director goal ('confirm environment features to navigate accurately') not present in the ground truth trace. The disfluency (uh) appears to trigger a spurious goal update. \\
\hline
\textit{``okay so u:m''} (t=3.31s) & Claude Sonnet 4 & 
None (correct) & 
No discrepancy flagged. LLM correctly withholds MM 
update for a turn consisting entirely of a filler token and 
elongated acknowledgment. \\
\hline
\end{tabular}
\caption{Representative examples from the human-validated 
discrepancy set for Dialogue 1, illustrating a spatial 
misgrounding failure, a disfluency-induced unsupported belief, 
and a correct trace. These examples support the pattern 
described in Section~6: LLMs generate linguistically plausible 
but environmentally ungrounded inferences, particularly in turns 
where spatial reference or transcription artifacts are present.}
\label{tab:qualitative}
\end{table*}

These examples illustrate the pattern described in Section 6: LLMs generate linguistically plausible but environmentally ungrounded inferences, particularly in turns where spatial reference or transcription artifacts are present.

\end{document}